\documentclass[lettersize,journal]{IEEEtran}
\usepackage{amsmath,amsfonts}
\usepackage{algorithm}
\usepackage{array}
\usepackage[caption=false,font=normalsize,labelfont=sf,textfont=sf]{subfig}
\usepackage{textcomp}
\usepackage{stfloats}
\usepackage{url}
\usepackage{verbatim}
\usepackage{graphicx}
\usepackage{cite}
\usepackage{bm}
\usepackage{enumitem}
\usepackage[noend]{algpseudocode}
\usepackage{color}
\usepackage{pifont}
\usepackage{wrapfig,lipsum,booktabs}
\algnewcommand\algorithmicinput{\textbf{Input:}}
\algnewcommand\Input{\item[\algorithmicinput]}
\algnewcommand\algorithmicoutput{\textbf{Output:}}
\algnewcommand\Output{\item[\algorithmicoutput]}

\usepackage[svgnames,table,xcdraw]{xcolor}
\definecolor{customred}{HTML}{C25759}
\definecolor{customblue}{HTML}{06007A}
\definecolor{customdarkred}{HTML}{7B4435}
\definecolor{customdarkgreen}{HTML}{447059}
\definecolor{customdarkyellow}{HTML}{CEA33F}
\definecolor{customedarkgray}{HTML}{595959}
\definecolor{customehighlight}{HTML}{ffffff}

\newcommand{\systemname}{\emph{AttributionScanner}}
\newcommand{\mosaicname}{\emph{Attribution Mosaic}}

\usepackage[most]{tcolorbox}

\usepackage{amssymb}
\newtcbox{\mybox}[1][gray]{on line, boxsep=2.5pt, boxrule=0pt, left=-1.2pt,right=-1.2pt,top=-1.2pt,bottom=-1.2pt, colback=gray, colframe=gray, arc=0.5mm}
\newtcbox{\myboxx}[1][gray]{on line, boxsep=2.5pt, boxrule=0pt, left=-0.5pt,right=-0.5pt,top=-1pt,bottom=-1pt, colback=gray, colframe=gray,arc=1.4mm}
\newtcbox{\myboxl}[1][Gainsboro]{on line, boxsep=2.5pt, boxrule=0pt, left=-1.2pt,right=-1.2pt,top=-1.2pt,bottom=-1.2pt, colback=Gainsboro, colframe=Gainsboro, arc=0.5mm}
\newtcbox{\myboxxA}[1][gray]{on line, boxsep=2.5pt, boxrule=0pt, left=-0.5pt,right=-0.5pt,top=-0.8pt,bottom=-0.8pt, colback=gray, colframe=gray,arc=1.622mm}  %%This is for text of sec 4
\newtcbox{\mylongbox}[1][gray]{on line, boxsep=2.5pt, boxrule=0pt, left=-1pt,right=-1pt,top=0.8pt,bottom=0.8pt, colback=gray, colframe=gray,arc=2.0mm}  %%This is for small of A1
\newtcbox{\mylongboxT}[1][gray]{on line, boxsep=2.1pt, boxrule=0pt, left=0pt,right=0pt,top=1pt,bottom=1pt, colback=gray, colframe=gray,arc=1.7mm}  %%This is for tniy of A1
\graphicspath{{figs/}{figures/}{pictures/}{images/}{./}} % where to search for the images

\usepackage{multirow}
\usepackage[colorlinks=true,linkcolor=customblue,citecolor=customblue,urlcolor=black]{hyperref}

\let\titleold\title
\renewcommand{\title}[1]{%
    \titleold{#1}% Call the original \title command
    \gdef\thetitle{#1}% Store the title globally for later use
}

\usepackage{soul}
\sethlcolor{yellow}

\begin{document}

\title{\systemname{}: A Visual Analytics System for Model Validation with Metadata-Free Slice Finding}
%%%%%%%%%%%%%%%%%%%%%%%%%%%%%%%%%%%%%%%%% \downarrow comment this out for appendix %%%%%%%%%%%%%%%%%%%%%%%%%%%%%%%%%%%%%%%%%
\author{Xiwei Xuan$^{1, \dag, *}$, Jorge Piazentin Ono$^2$, Liang Gou$^{3, \dag}$, Kwan-Liu Ma$^1$, and Liu Ren$^2$
        % <-this % stops a space
\thanks{$^1$ Department of Computer Science, University of California, Davis, CA 95616, USA. E-mails: \{xwxuan, klma\}@ucdavis.edu.}
        % <-this % stops a space
\thanks{$^2$ Bosch Center for Artificial Intelligence (BCAI), Bosch Research North America. E-mails: \{jorge.piazentinono, liu.ren\}@us.bosch.com.}
        % <-this % stops a space
\thanks{$^3$ Splunk Technology, San Jose, CA, USA. E-mail: lgou.psu@gmail.com.}
        % <-this % stops a space
\thanks{$\dag$ This work was done when the authors worked with Bosch Research North America.}
        % <-this % stops a space
\thanks{$*$ Corresponding author. E-mail: xwxuan@ucdavis.edu.}
}

\markboth{Journal of \LaTeX\ Class Files,~Vol.~14, No.~8, August~2021}%
{Shell \MakeLowercase{\textit{et al.}}: A Sample Article Using IEEEtran.cls for IEEE Journals}

\maketitle


\begin{abstract}
Data slice finding is an emerging technique for validating machine learning (ML) models by identifying and analyzing subgroups in a dataset that exhibit poor performance, often characterized by distinct feature sets or descriptive metadata. However, in the context of validating vision models involving unstructured image data, this approach faces significant challenges, including the laborious and costly requirement for additional metadata and the complex task of interpreting the root causes of underperformance. To address these challenges, we introduce \systemname{}, an innovative human-in-the-loop Visual Analytics (VA) system, designed for metadata-free data slice finding. Our system identifies interpretable data slices that involve common model behaviors and visualizes these patterns through an \mosaicname{} design. Our interactive interface provides straightforward guidance for users to detect, interpret, and annotate predominant model issues, such as spurious correlations (model biases) and mislabeled data, with minimal effort. Additionally, it employs a cutting-edge model regularization technique to mitigate the detected issues and enhance the model's performance. The efficacy of \systemname{} is demonstrated through use cases involving two benchmark datasets, with qualitative and quantitative evaluations showcasing its substantial effectiveness in vision model validation, ultimately leading to more reliable and accurate models.
 
\end{abstract}


\IEEEpubidadjcol

\section{Introduction}
\label{sec:intro}
As machine learning (ML) technologies continue to prevail in various domains, the importance of model interpretation and validation has grown significantly. Detecting a model's failure and understanding the underlying reasons are crucial in promoting greater model accuracy, transparency, and accountability. 
A prevalent and harmful issue in ML models is ``spurious correlation''~\cite{simon1954spurious}, where a model erroneously utilizes irrelevant image features for classification. For instance, in hair color classification, a model should make decisions according to hair features, termed as ``core correlation'', while the use of other features (e.g., background or facial attributes) denotes ``spurious correlation.'' Such spurious correlations, regardless of model accuracy, pose substantial challenges including poor generalization in production~\cite{singla2022core} and AI fairness issues~\cite{madaio2020co}.

However, models often perform inconsistently across different subsets of data, making it challenging to validate a model comprehensively to uncover hidden issues~\cite{sagawa2019distributionally,zhang2022correct,liu2021just,xuan2024slim,kirichenko2022last}. Thus, the crucial demands of identifying and characterizing such problematic data subsets have fueled data slice finding techniques~\cite{sagawa2019distributionally,zhang2022correct,liu2021just,kirichenko2022last,pastor2021looking,zhang2022sliceteller}. The term ``data slice'' refers to data subgroups with coherent features, often defined by metadata, model representations, or other descriptive information. By detecting, understanding, and resolving issues in data slices, ML experts can improve model performance and ensure safer and more reliable deployments in high-stakes domains~\cite{foster2011subgroup,zhang2018deeproad,farchi2021ranking,eyuboglu2022domino}.
To enhance the trustworthiness, eXplainable Artificial Intelligence (XAI) techniques~\cite{gunning2017explainable} are also utilized to reveal potential factors adversely impacting the model and apply targeted model regularization~\cite{rieger2020interpretations}.

Despite the promising benefits of these techniques, several hurdles remain unaddressed for a thorough model validation.

\noindent\textbf{Challenges of Data Slice Finding.} 
State-of-the-art methods for data slice finding demand extensive metadata associated with each single instance, which are often obtained from manual annotations~\cite{chung2019slice,sagadeeva2021sliceline,pastor2021divergent} or generated by pre-trained vision-language models~\cite{eyuboglu2022domino}. However, it is expensive to collect manually annotated metadata, and vision-language models can produce erroneous results~\cite{xuan2025vista}, especially in domain-specific scenarios. When neither metadata nor suitable pre-trained models are available, obtaining data slices becomes challenging: requires manual examination of numerous instances or computationally expensive model training~\cite{sohoni2020no,deon2022spotlight}. 

\noindent\textbf{Challenges of XAI for Vision Models}.
Explanation methods for vision models are often in instance-level~\cite{ribeiro2016should,zhou2016learning,selvaraju2017grad,petsiuk2018rise}, producing explanations for individual data instances and leading to less-actionable insights---When observing a problematic model behavior, it is unknown whether it is a rare or common case, whether an actual model issue happens is still unclear. To obtain more reasonable conclusions, a thorough examination of all instances is required yet practically unfeasible.

\noindent\textbf{Our approach.} We present \systemname{}, a Visual Analytics (VA) system designed for efficient vision model validation with slice finding, eliminating the costly requirements of metadata or other descriptive information. Our innovative workflow employs attribution-based explanation techniques to create interpretable feature representations for slice finding, enabling discriminative slice construction and summary to streamline user exploration.
Additionally, the methods provided by \systemname{} are applicable to various computer vision models, collaboratively forming a human-in-the-loop approach for model validation and enhancement.
Our main contributions include: 
\begin{enumerate}[label=\textbullet, align=left, leftmargin=!
]
    \item \systemname{}, a novel framework and VA system to validate vision models with data slice finding. Unlike existing methods, \systemname{} removes the resource-intensive need for metadata to generate interpretable data slices. 
    
    \item A design for slice summarization, \mosaicname{}, illustrating the significant pattern of model behaviors in data slices, which reduces the effort in conventional approaches of examining a large volume of individual data samples. 

    \item An iterative workflow supporting model validation and enhancement, which ensures generated insights are actionable, enabling users to swiftly detect, understand, and fix model issues related to spurious correlations and mislabeled data.
    
\end{enumerate}

% This paper is structured as follows: in Sec.~\ref{sec:rw}, we review related work on data slice finding and XAI that is relevant to our work Sec.~\ref{sec:methodology} outlines the methodology and system design of \systemname{}. In Sec.~\ref{sec:usecases}, we illustrate two use cases showcasing how \systemname{} can be utilized for vision model validation. An evaluation of our system and experts' feedback are discussed in Sec.~\ref{sec:evaluation}. Lastly, we present our future work and conclusions in Sec.~\ref{sec:conclusion}.

\section{Related Work}
\label{sec:rw}
% Our method, \systemname{}, leverages insights from both Data Slice Finding and XAI techniques to pinpoint problematic data slices in vision-based models. This provides a brief overview of the relevant literature on these two topics.

\subsection{Data Slice Finding} \label{subsec:rw_slice_finding}
Data slice finding is a critical technique for identifying coherent subsets of data where ML models may perform inconsistently. These subsets, or slices, are often defined by similar metadata, model representations, data statistics, or other descriptive information~\cite{barash2019bridging,chung2019slice,pastor2021looking,sagadeeva2021sliceline}. By analyzing problematic data slices, ML experts can design targeted solutions to address model issues in edge cases, improving the model robustness and reliability~\cite{sagawa2019distributionally,xuan2024slim}. However, existing approaches often rely on expensive metadata to obtain meaningful subgroups, limiting their applicability. To address this, recent work has focused on leveraging feature vectors in the latent space and clustering them to produce data slices~\cite{sohoni2020no,deon2022spotlight,krishnakumar2021udis,lee2022viscuit}. For example, Domino~\cite{eyuboglu2022domino} uses a pre-trained vision-language model (VLM) to generate textual descriptions of slices. However, this approach can backfire if there is a significant domain mismatch, yielding less meaningful slices. Our work fills this gap by introducing a system that provides explainable slices without requiring metadata or pre-trained VLMs, making it more broadly applicable and reliable.

\subsection{Visual Analytics for Model Explanation and Validation} \label{subsec:rw_model_validation}
Visual analytics has emerged as a powerful tool for both model explanation~\cite{xuan2022vac,lai2023explore, wang2024visual,prasad2022transform,huang2022conceptexplainer,chen2020oodanalyzer,li2022unified} and validation~\cite{robertson2023angler,chen2023unified,zhang2024slicing,yang2024foundation,jin2023shortcutlens,zhang2023labelvizier}. For instance, Kahng et al.~\cite{kahng2016visual} use pairwise feature combinations to produce data slices and assess model performance. FairVis~\cite{cabrera2019fairvis} enables domain users to manually slice data to identify model biases. Similarly, Errudite~\cite{wu2019errudite} develops a domain-specific language for slicing textual documents. While manual slicing is useful, it is not scalable due to its labor-intensive nature. To address this, SliceTeller~\cite{zhang2022sliceteller} combines an automatic slice-finding algorithm with a visual analytics tool, allowing for iterative refinement of models. However, it still relies on structured metadata to produce meaningful slices. Our approach, \systemname{}, eliminates the need for expensive metadata while incorporating effective human participation, making it a scalable and accessible solution for explainable model validation.

\begin{figure*}
  \centering
  \includegraphics[width=0.9\linewidth]{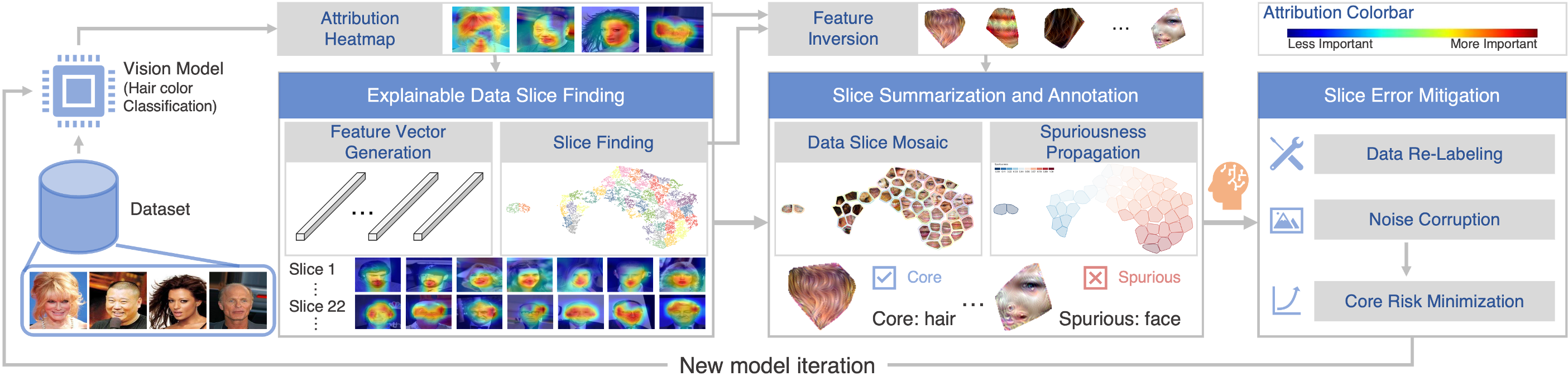}
  \caption{\systemname{} workflow involves three phases: \emph{Explainable Data Slice Finding}, \emph{Slice Summarization and Annotation}, and \emph{Slice Error Mitigation}. The first phase: GradCAM is used to assist the generation of feature vectors and then obtain data slices. The second phase: users can identify and annotate slice error types such as core/spurious correlations and noisy labels with the help of \mosaicname{} and Spuriousness propagation. The third phase: the annotation and user-verified Spuriousness are used on the ML side to mitigate slice errors.
  % including (1) Data re-labeling if considered necessary, and (2) noise corruption and Core Risk Minimization (CoRM) applied if spurious correlations are detected.
  }
  \label{fig:workflow}
\end{figure*}

\subsection{XAI Techniques for Neural Networks} \label{subsec:rw_xai}
Explainable AI (XAI) techniques have become essential for interpreting neural networks, often using visualizations to provide insights into model behavior. Attribution-based methods~\cite{zhou2016learning,selvaraju2017grad,ribeiro2016should,xuan2024suny,petsiuk2018rise}, such as GradCAM~\cite{selvaraju2017grad}, generate heatmaps or decision boundary explanations for individual instances. While these methods offer intuitive and straightforward results, they are instance-based and fail to capture broader patterns in model behavior. Another line of work utilizes optimization to produce synthetic images for model explanation, such as Feature Visualization~\cite{erhan2009visualizing,olah2017feature} and Feature Inversion~\cite{mahendran2015understanding}, which create visualizations that maximally activate specific neurons/layers, or reconstruct images from feature vectors. These methods provide insights into the learned patterns of a model but are often limited to single instances or randomly aggregated groups. In \systemname{}, we bridge this gap by leveraging Feature Inversion to highlight shared patterns across data slices, providing visual summaries of model behavior at the slice level. Additionally, we integrate GradCAM to explain individual instances, offering a comprehensive view of model behavior at both the micro and macro levels.

\section{\systemname{}} 
\label{sec:methodology}
% In this section, we introduce \systemname{}, a novel approach for interpretable ML model validation with metadata-free, data slice-level analysis. To foster a thorough understanding of our system, we first outline the requirements that shaped our design choices. Then, we delve into the visual components of \systemname{} that enable interpretable model validation. We then detail our slice computation and summarization method, which allow us to generate data slices with consistent model attribution without additional metadata. Finally, we show how we leverage the insights obtained from our system to enhance the model.

\subsection{Design Requirements} \label{subsec:design_requirement}

Slice finding has garnered increased attention due to its potential to facilitate a thorough ML model evaluation. However, this task is particularly challenging in the realm of vision models considering the inherent unstructured nature of vision data. A pivotal shortcoming we observed in existing approaches is their dependency on metadata, including pre-collected or vision-language model-generated ones, to compute meaningful data slices. However, the acquisition of metadata is resource-intensive, demanding substantial human effort and financial investment. Additionally, the vision-language models are commonly trained on general-purpose datasets, potentially leading to inaccurate or biased results in specific domains. Stemming from these observations, we delineated the subsequent requirements for a system capable of conducting metadata-free, slice-driven model validation:

\begin{enumerate}[label={\bfseries R\arabic*.}, ref={\color{customblue} \bfseries R\arabic*}, start=1]
  \item \label{req:noprior} \emph{Metadata-free}. Our method should yield meaningful data slices without metadata, i.e.,  neither using metadata as input nor generating or extracting metadata.
  \item \label{req:interpretability} \emph{Interpretability}. We should provide sufficient and intuitive information to enhance model transparency, supporting interpretations of a data slice's shared attributes for streamlined exploration and individual instances for insight verification as needed.
  \item \label{req:summary} \emph{Slice Overview}. Mandatory manual inspection of individual images is impractical for model validation, especially with a high volume of data.
  To enhance efficiency, our system should provide a visual summary of data slices, depicting each data slice's key features and attributions.
  \item \label{req:actionable} \emph{Actionable insights}. The system should ensure the obtained insights are actionable, which requires careful design of visual guidance, issue annotation, and corresponding model or data regularization.
\end{enumerate}

\subsection{System Workflow} \label{subsec:system_workflow}

\systemname{} is a human-in-the-loop system, taking as input an image dataset and a trained CNN, such as ResNet~\cite{he2016deep} for image classification, and yielding interpretable data slices to assist experts in efficient model validation.

A workflow of our approach is presented in Fig.~\ref{fig:workflow}, where we foster explainable slice finding without metadata (\ref{req:noprior}, \ref{req:interpretability}) by leveraging GradCAM~\cite{selvaraju2017grad} and Feature Inversion~\cite{olah2017feature}. The first ``explainable data slice finding'' phase interpolates model attributions into the latent space to craft attribution-weighted feature vectors, forming an attribution representation space with locally consistent model attributes (\ref{req:noprior}). Then we conduct clustering over this space to generate data slices. Although the produced slices maintain consistent attributions, they still demand significant human effort to identify and understand their shared patterns. 
To streamline slice exploration and issue detection, the ``slice summarization and annotation'' phase introduces \mosaicname{}, which visually elucidates each slice's dominant attributions (\ref{req:summary}), synchronized with instance-level heatmaps for insight verification (\ref{req:interpretability}). 
Upon user annotation of an issue like ``spurious correlations'', our Spuriousness Propagation method automatically estimates a spuriousness score for each slice to help users uncover other hidden model biases (\ref{req:actionable}). Afterward, our workflow enables the mitigation of detected issues in data or model, corresponding to the ``slice error mitigation'' phase (\ref{req:actionable}).

% The rest of this section delineates our interface for data slicing-based model validation, explains our presented data slicing and summarization approach, discusses design considerations, and demonstrates how the model can be fine-tuned to rectify detected slice issues.

\begin{figure*}[t!]
\centerline{\includegraphics[width=0.9\linewidth]{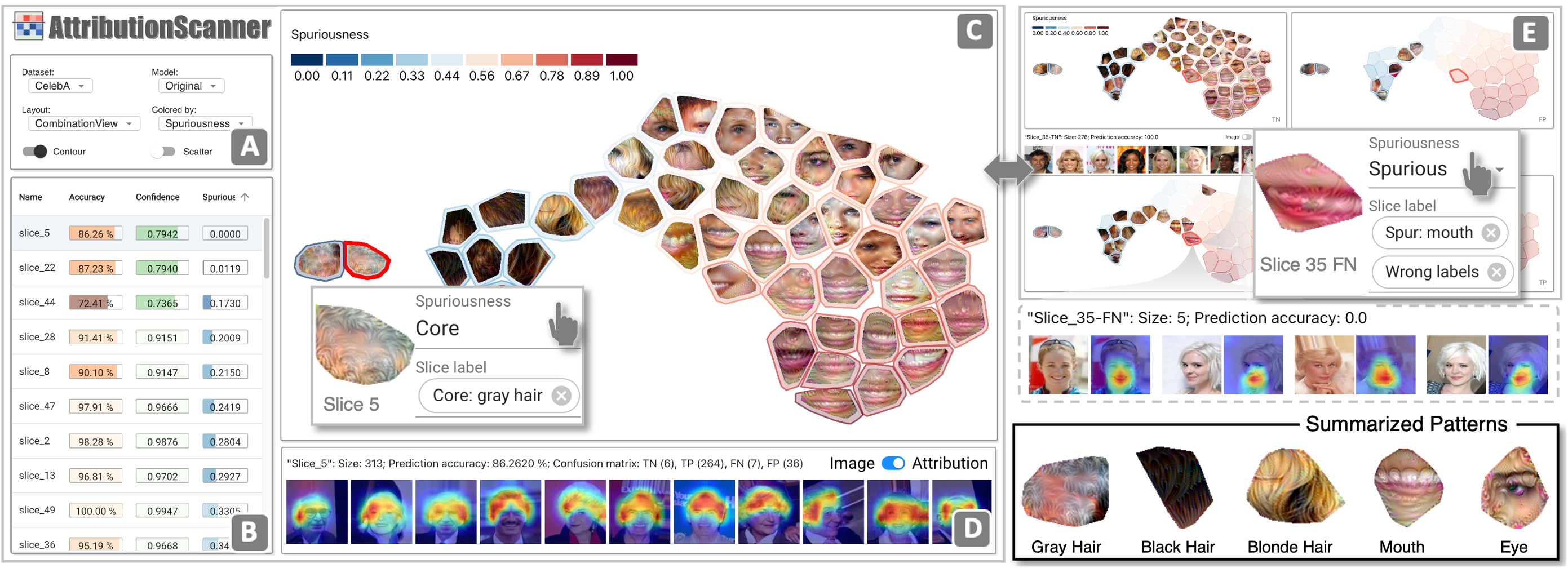}}
\caption{
\systemname{} applied to the model validation of a hair color classifier trained on the CelebA dataset. (A) System Menu, enabling the selection of dataset, model, and visualization options (Layout, Confusion Matrix, Scatter Plot/\mosaicname{}, and Contour Visibility). (B) Slice Table, showing slice metrics. (C) \mosaicname{}, showing a visual overview of all data slices, which can also be displayed as a confusion matrix view (E). (D) Slice Detail View, showing individual images or Attribution Heatmaps belonging to a selected data slice.
}\label{fig:teaser}
\end{figure*}

\subsection{System Interface} \label{subsec:system_interface}

\systemname{} is comprised of four main components. Fig.~\ref{fig:teaser} depicts a use case where our system is used to validate a hair color classification model.

The System Menu (Fig.~\ref{fig:teaser}~\mybox{\textcolor{white}{A}}) provides options for dataset and model selection, visualization layout configuration, and color encoding choice. Two visualization layouts are available: Combination view and Confusion Matrix view, allowing for an overall view of data slices or a more detailed inspection categorized by error types (\ref{req:summary}). The provided color encoding choices include Slice Name, Slice Accuracy, Slice Confidence, and Spuriousness Probability, adaptable to user requirements.

The Slice Table (Fig.~\ref{fig:teaser}~\mybox{\textcolor{white}{B}}) presents various slice performance metrics to facilitate dataset navigation. Available metrics include Accuracy, Confidence, and Spuriousness Probability, which is produced by the Spuriousness Propagation method upon user annotation (Sec.~\ref{subsec:slicesummarization}). This table assists in detecting intriguing patterns, like high accuracy with wrong attributions possibly indicative of model biases. On the other hand, low accuracy with correct attributions indicates the potential existence of mislabeled data (\ref{req:summary}, \ref{req:actionable}).

Our system's third component is the \mosaicname{}, rendering the dominant visual patterns of each slice with respect to the model's decision (\ref{req:interpretability}, \ref{req:summary}), either in a unified form (Fig.~\ref{fig:teaser}~\mybox{\textcolor{white}{C}}) or segregated by the model's confusion matrix (Fig.~\ref{fig:teaser}~\mybox{\textcolor{white}{E}}). The design method for this view is detailed in Sec.~\ref{subsec:slicesummarization}. Moreover, \mosaicname{} incorporates visualization of user-specified metrics through the coloration of mosaic boundaries, providing a vital guide in pinpointing problematic slices (\ref{req:summary}). Annotation of a slice is facilitated by a double-click action on a mosaic tile, allowing for effortless insight collection (\ref{req:actionable}).

The Slice Detail View (Fig.~\ref{fig:teaser}~\mybox{\textcolor{white}{D}}) showcases individual image samples of the selected data slice, rendering either images or attribution heatmaps. This view enables an in-depth examination of each slice (\ref{req:interpretability}).

\begin{figure}
  \centering\includegraphics[width=\linewidth]{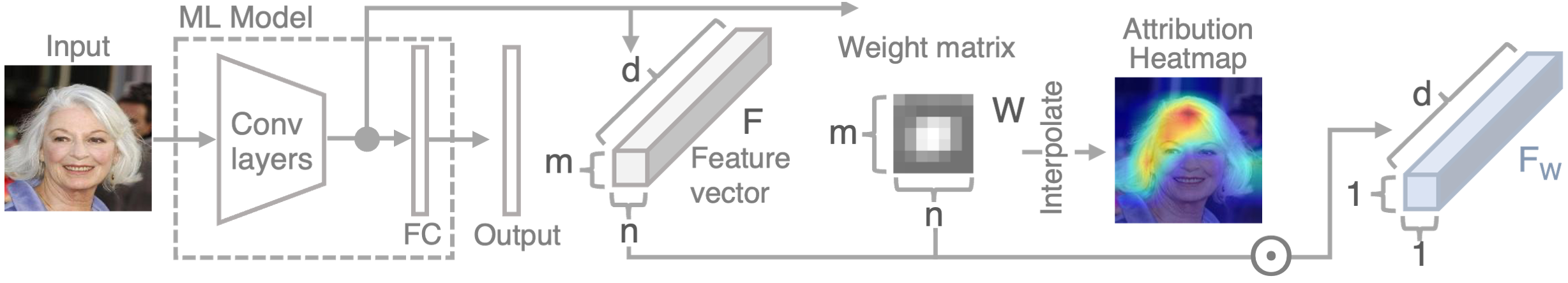}
  \caption{Attribution-weighted feature vector generation. An image is forwarded through CNN, where the corresponding feature vector $F$ and weight matrix $W$ can be extracted to calculate the attribution-weighted feature vector $F_W$.
  }
  \label{fig:GradCAMfeatureExtraction}
\end{figure}

\subsection{Explainable Data Slice Finding} \label{subsec:exp_slice_finding}

In this section, we delineate our Explainable Data Slicing method. Our approach presumes a typical CNN model architecture, comprising a sequence of convolutional layers succeeded by a fully connected (FC) layer, a commonality across most CNN models~\cite{simonyan2014very, he2016deep, selvaraju2017grad}. 

% The process unfolds in three phases: initially, we create attribution-weighted feature vectors utilizing attribution masks in the latent space. Subsequently, we derive an attribution representation space and compute data slices. Lastly, we employ the Feature Inversion~\cite{mahendran2015understanding} to distill shared visual patterns per slice and visualize these patterns alongside model performance indicators in \mosaicname{}.

\subsubsection{Attribution-Weighted Feature Vector Generation}
In this section, we detail the generation of attribution-weighted feature vectors (\ref{req:noprior}, \ref{req:interpretability}).
It is notable that there are existing approaches employing attribution-weighted rather than original feature vectors, which aim to provide better model explanation~\cite{mahendran2015understanding,sikdar2021integrated,janizek2021explaining} or conduct explainable model enhancement~\cite{rieger2020interpretations,xuan2024slim}. Considering attribution-weighted feature vectors reserve both data features and model attributes~\cite{janizek2021explaining,rieger2020interpretations,xuan2024slim}, we opt for this design choice to support our explainable data slice construction. Specifically, we leverages GradCAM to obtain model attributes in the latent space, due to its demonstrated efficacy via sanity checks~\cite{adebayo2018sanity} and prevalent usage~\cite{rieger2020interpretations,krishnakumar2021udis,mao2022causal,zhuang2019care,yang2023mitigating,wu2023discover,leetowards}. In our approach, GradCAM is interchangeable with other XAI techniques such as CAM~\cite{zhou2016learning} or RISE~\cite{petsiuk2018rise} (\ref{req:interpretability}).

As shown in Fig.~\ref{fig:GradCAMfeatureExtraction}, an input image is processed by the model to extract the feature vector $F$, and GradCAMis performed to produce attributes in the latent space $W$, which can be linearly interpolated to the image size for attribution heatmap. $W \in \mathbb{R}^{m\times n}$ and is normalized with $\Sigma_{i=1,j=1}^{i=m, j=n} W_{ij}=1$ to compute the weighted average of $F$. The resulting attribution-weighted feature vector $F_W \in \mathbb{R} ^{1\times 1 \times d}$ is:
\begin{equation}\label{eqn:eval1}
F_W = F \odot W = \Sigma_{i=1,j=1}^{i=m, j=n} F_{ij} W_{ij}.
\end{equation}
% $$
% F_W = \frac{\Sigma_{i=1,j=1}^{i=m, j=n} F_{ij} W_{ij}}{\Sigma_{i=1,j=1}^{i=m, j=n} W_{ij}}
% $$

\begin{figure}
  \centering\includegraphics[width=\linewidth]{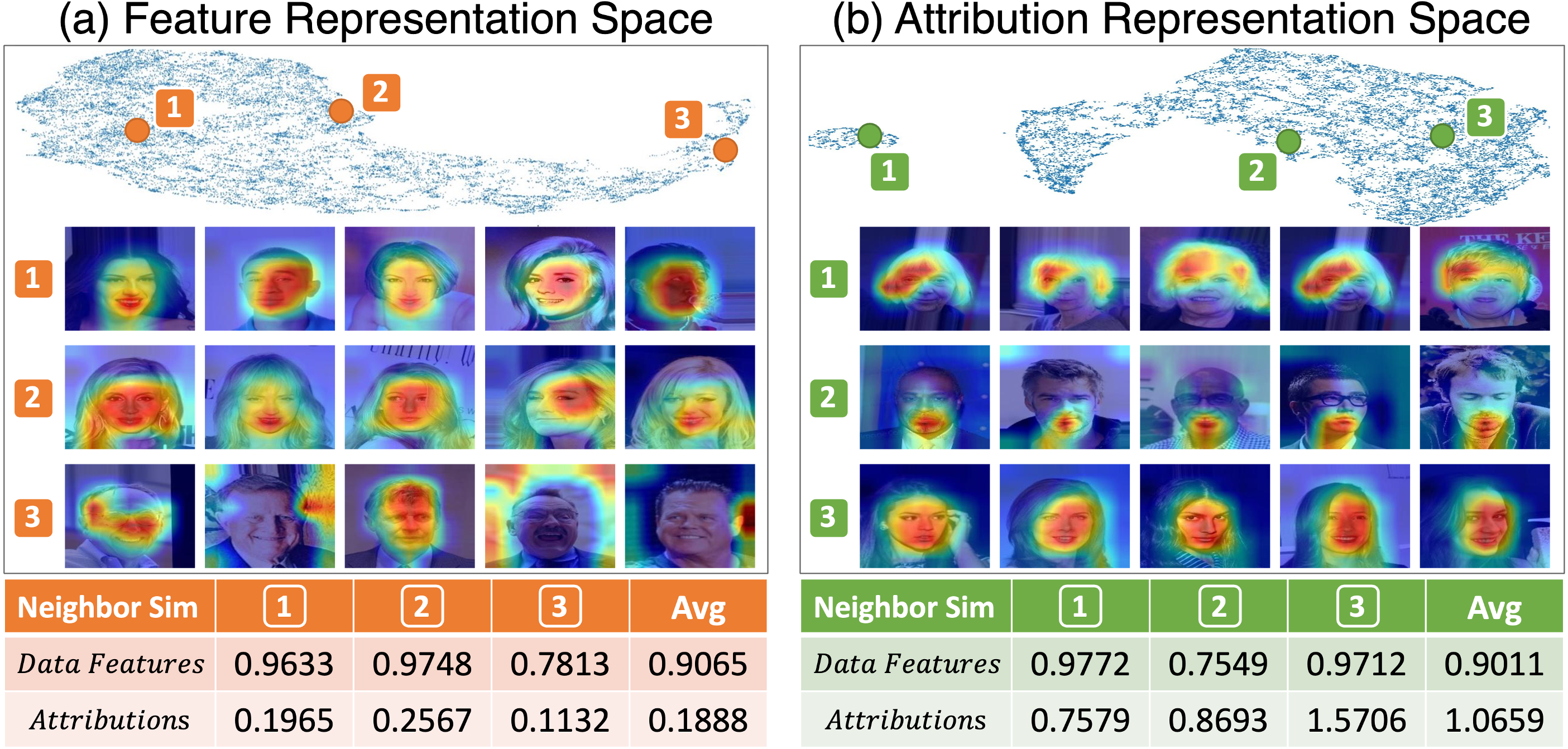}
  \caption{Comparison of representation spaces. (a) Feature representation space computed on original feature vectors. (b) Attribution representation space computed on attribution-weighted feature vectors. 
  % Highlighted images indicate that only (b) maintains neighbor consistency regarding similar model attributions.
  }
  \label{fig:proj_comparison}
\end{figure}

\subsubsection{Data Slice Identification}

\textbf{Representation Space Construction.} 
To derive meaningful data slices (\ref{req:noprior}, \ref{req:interpretability}), we need to construct a representation space with neighbor consistency in terms of model attributes and data features. We first conduct quantitative and qualitative experiments to validate whether attribution-weighted feature vectors help fulfill this requirement---by comparing the feature representation space (derived from the original feature vectors) and the attribution representation space (derived from the attribution-weighted feature vectors).
Following the established practices~\cite{eyuboglu2022domino,sagawa2019distributionally,zhang2022correct} and considering the time efficiency~\cite{mcinnes2018umap,pal2020performance}, we utilize dimensionality reduction by UMAP~\cite{mcinnes2018umap} to produce both spaces. At each space, three points are randomly sampled and their nearest neighbors are obtained. We qualitatively examine the consistency of their corresponding attribution heatmaps as shown in Fig.~\ref{fig:proj_comparison}. Neighboring instances' model attributions on the original space are not similar, while our constructed space has greater local consistency of model attributions.

In addition, we quantitatively compute the neighboring consistency regarding data features and model attributions and report the results in Fig.~\ref{fig:proj_comparison}. For each point, we compute the average cosine similarities between its top 10 neighbors using their original feature vectors, representing data feature consistency; besides, we compute the average Wasserstein similarities of their attribution mask, where higher values of both scores indicate better consistency. Lastly, we compute the average scores to quantify the overall neighbor consistency of data features and model attributions for each space. As shown in Fig.~\ref{fig:proj_comparison}, compared to the original feature space, our constructed space has comparable data feature similarity (0.8987) and significantly higher attribution similarity (1.0659), which proves that the attribution-weighted feature vectors help construct a space with satisfactory local consistency in terms of data features and model attributions, providing a solid basis for computing interpretable data slices (\ref{req:interpretability}).

\textbf{Slice Identification.} 
On top of the attribution representation space, we apply K-Means clustering to obtain data slices. Similarly to Domino~\cite{eyuboglu2022domino}, we apply over-clustering by incrementing the number of clusters and monitoring the model attribution consistency within each slice, until a coherent grouping of samples is attained. Specifically, the attribution consistency of a data slice is measured as the average cosine similarity of each instance's attribution-weighted feature vector with the slice centroid. We set a threshold at $0.8$ to determine slice coherence in our case studies.
\begin{figure}[!t]
  \centering
  \includegraphics[width=\linewidth]{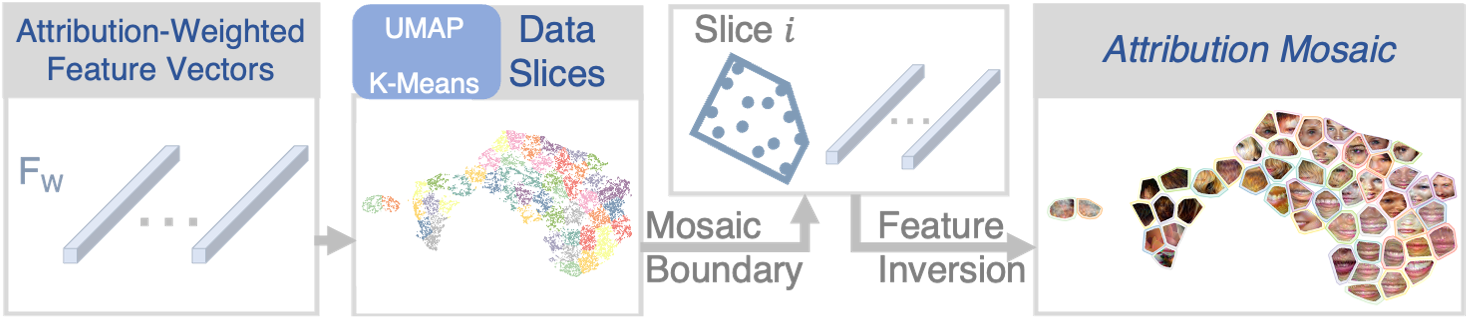}
  \caption{\mosaicname{} generation. We compute data slices after acquiring attribution-weighted feature vectors ($F_W$). Then Feature Inversion is conducted according to $F_W$ and the mosaic boundary of each slice to visualize their common patterns, forming \mosaicname{}.}
  \label{fig:sliceSummarization}
\end{figure}
Note that we allow ML experts to adjust this threshold based on the specific requirements of their datasets and tasks. For rare slices with a small number of data samples featuring abnormal model attributions, this iterative process, along with our slice mosaic design detailed in Sec.~\ref{subsec:slicesummarization}, ensures such slices are not overlooked. By combining automated clustering with user-driven refinement, our approach provides a comprehensive and efficient solution for identifying both common and rare slices.

\subsection{Slice Summarization and Annotation}\label{subsec:slicesummarization}

\subsubsection{Attribution Mosaic}

After obtaining slices with internally consistent model attributions and covering the entire dataset, we design~\mosaicname{}, a novel visualization technique, to visually summarize data slices in a mosaic depiction(\ref{req:summary}). As shown in Fig.~\ref{fig:sliceSummarization}, we first compute the convex hull of each slice based on the projected points in 2D space; then distill each slice's predominant visual patterns via Feature Inversion, integrating the shape constraint of the convex hull. The final landscape featuring slices' attribution summaries is presented as~\mosaicname{}, allowing users to easily identify slices with rare, or uncommon model attributions through visual inspection and comparison.

\textbf{Mosaic Boundary Generation.} 
Given that the convolutional modules take rectangular-shaped input, conventional Feature Inversion produces rectangular visualizations, merely obtaining and displaying its results over the projected cluster centroids leads to overlapped visualizations, hindering effective user exploration. Since K-Means is a centroid-based algorithm that partitions the space into strict non-overlapping Voronoi cells, this guarantees that the convex hulls~\cite{barber1996quickhull} of our data slices have no overlap.
Therefore, we compute the convex hull of each slice on the attribution representation space, which provides a mosaic layout for our slice visualization.

\textbf{Mosaic Drawing with Feature Inversion.}
We deploy Feature Inversion technique~\cite{carter2019activation,nguyen2016multifaceted,fel2023unlocking} to visualize the significant pattern in each slice (\ref{req:summary}), which is an optimization-based technique to visualize shared features across a group of feature vectors.
Specifically, we first create an image $I$ with size $W \times H$, which is the minimal rectangle encompassing the mosaic boundary of a specific slice denoted by $M$. The optimization process focuses on updating pixels in $I$ that lie inside $M$, synthesizing a mosaic-shaped image to visualize the dominant pattern of the feature vectors. To enhance the quality of synthesized images, we initialize them with the average of the original images instead of Gaussian noises~\cite{nguyen2016multifaceted} and parameterize them in the Fourier space~\cite{fel2023unlocking}, making them better aligned with natural images rather than less-interpretable artificial patterns.
The optimization process is defined by:
\begin{equation}
   x^* = \operatorname{arg\,max}_{x \in M} \left(\ell(\phi(x), \phi_0) + \lambda R(x) \right), 
\end{equation}
\begin{table}[!t]
\caption{System Design Considerations.}
\label{tab:design_consideration}
\resizebox{\columnwidth}{!}{%
\begin{tabular}{cccc|cccc}
% \rowcolor[HTML]{5F97D2}
\multicolumn{4}{c|}{\cellcolor[HTML]{5F97D2}{\color{white} \textbf{Design Components}}} & \multicolumn{4}{c}{\cellcolor[HTML]{8066B4}{\color{white} \textbf{System Capabilities}}}\\
\cellcolor[HTML]{8EB8F5}{\color{white} \begin{tabular}[c]{@{}c@{}}Slice \\[-0.1pt] Table\end{tabular}} & \cellcolor[HTML]{8EB8F5}{\color{white} \begin{tabular}[c]{@{}c@{}}Slice Detail \\[-0.1pt] View\end{tabular}} & \cellcolor[HTML]{8EB8F5}{\color{white} \begin{tabular}[c]{@{}c@{}}Scatter \\[-0.1pt] Plot\end{tabular}} & \cellcolor[HTML]{8EB8F5}{\color{white} \begin{tabular}[c]{@{}c@{}}Attribution\\[-0.1pt] Mosaic\end{tabular}} & \cellcolor[HTML]{A080E1}{\color{white} \begin{tabular}[c]{@{}c@{}}Slice \\[-0.1pt] Performance\end{tabular}} & \cellcolor[HTML]{A080E1}{\color{white} \begin{tabular}[c]{@{}c@{}}Slice \\[-0.1pt] Attribution\end{tabular}} & \cellcolor[HTML]{A080E1}{\color{white} \begin{tabular}[c]{@{}c@{}}Instance \\[-0.1pt] Performance\end{tabular}} & \cellcolor[HTML]{A080E1}{\color{white} \begin{tabular}[c]{@{}c@{}}Instance \\[-0.1pt] Attribution\end{tabular}} \\
\rowcolor[HTML]{E6F0FF}
\checkmark &\checkmark   &   &   & \cellcolor[HTML]{EEE6F7} \checkmark  & \cellcolor[HTML]{EEE6F7} & \cellcolor[HTML]{EEE6F7} & \cellcolor[HTML]{EEE6F7} \checkmark   \\
\rowcolor[HTML]{F8FAFE} \checkmark  &  \checkmark  &  \checkmark  &                                      & \cellcolor[HTML]{FAF5FF} \checkmark                                       & \cellcolor[HTML]{FAF5FF}                                                                 & \cellcolor[HTML]{FAF5FF}  \checkmark                                                     & \cellcolor[HTML]{FAF5FF}  \checkmark                                                                                     \\
\rowcolor[HTML]{E6F0FF} \checkmark & \checkmark    &   &   \checkmark                       & \cellcolor[HTML]{EEE6F7}  \checkmark                                                     & \cellcolor[HTML]{EEE6F7}  \checkmark                                                     & \cellcolor[HTML]{EEE6F7}  \checkmark                                                    & \cellcolor[HTML]{EEE6F7} \checkmark                                                                                      
\end{tabular}%
}
\end{table}
where $x^*$ denotes the pixels in $I$ bounded by the mosaic shape $M$, the convex hull of the slice. The values of $x^*$ are updated iteratively through the optimization process to maximize the target function. $\phi: \mathbb{R}^{W \times H \times C} \rightarrow \mathbb{R}^d$ is the representation function, i.e., the model's convolutional modules. $\phi_0$ is a target feature activation, which is the average of attribution-weighted feature vectors of a data slice in our case. $\ell(*)$ is the loss function where the Euclidean Loss is used here, and $R(x)$ is a Total Variation (TV) regularization term to improve the quality of results~\cite{mahendran2015understanding}.
Thus, we obtain a mosaic-shaped image to visualize the shared attribution patterns within a data slice. Lastly, the visualizations from all data slices are exhibited at their respective positions within the \mosaicname{}.

\subsubsection{Slice Annotation and Spuriousness Propagation}

The \mosaicname{} facilitates rapid comprehension of the main content of different slices, supporting the identification of potential spurious correlation issues(\ref{req:actionable}).

We introduce a metric termed Spuriousness probabilities to streamline the detection of spuriousness, ranging from $0$ to $1$, indicative of the likelihood of spurious correlation existence. We employ the Label Spreading algorithm~\cite{zhou2003learning} to propagate user-annotated spurious/core information to other slices.
Specifically, a graph is constructed where nodes represent data points, and edge weights encode similarities between points. In our case, we fit the graph to our 2D attribution representation space.
Based on this graph structure, the Label Spreading algorithm iteratively updates the labels of unannotated slices, ensuring local and global label consistency—similar slices should exhibit similar spuriousness scores, and the propagated scores should respect the annotated slice spurious/core labels.
While only one annotation of ``core'' or ``spurious'' is required to initialize the propagation process, the spuriousness scores are dynamically updated as users provide additional annotations. The iteratively refined scores are visualized in the interface to guide further annotation. 

The Spuriousness probabilities obtained through the \mosaicname{} offer two benefits. First, they streamline slice exploration by highlighting potential spurious correlations, supporting identifying and assessing problematic slices. Second, after users' verification, these probabilities inform the model-side strategies for issue mitigation (Sec.~\ref{subsec:error_mitigation}) (\ref{req:actionable}). Since propagated scores are visualized alongside the \mosaicname{}, users can easily validate and override incorrect propagations, ensuring human decisions take precedence over automated computations.

\begin{figure}[!t]
  \centering
  \includegraphics[width=.8\linewidth]{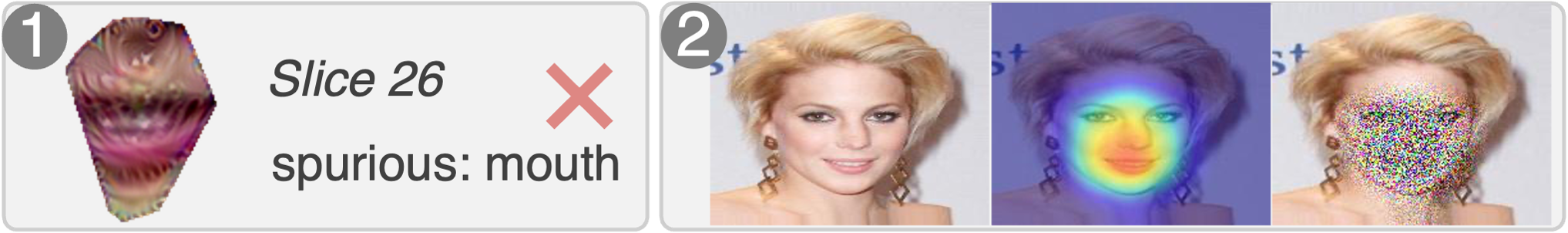}
  \caption{Example of spurious correlation mitigation with CoRM for hair color classification: \myboxx{\textcolor{white}{1}} The \mosaicname{} indicates a problem with a spurious feature: mouth. \myboxx{\textcolor{white}{2}} An example of how noise is occupied by CoRM: Original Image (left), GradCAM activations (middle), noise added to the spurious regions of the image (right).
  % For illustration purposes, the gaussian noise added is exaggerated.
  % \myboxx{\textcolor{white}{3}} More examples with the gaussian noise added to the spurious regions.
  }
  \label{fig:spurious_CoRM}
\end{figure}

\subsection{System Design Considerations} 
\label{subsec:design_consideration}

In designing \systemname{}, various configurations of visual components' combinations have been evaluated to ensure optimal system capabilities to fulfill our design requirements discussed in Sec.~\ref{subsec:design_requirement}, as outlined in Table~\ref{tab:design_consideration}.
Our system should be capable of providing four-faced information: (1) slice performance indicator (\ref{req:summary}), (2) slice attribution overview (\ref{req:interpretability}, \ref{req:summary}), (3) instance performance indicator (\ref{req:interpretability}), and (4) instance attribution inspection (\ref{req:interpretability}, \ref{req:actionable}). The notions in ``design components'' are consistent with Fig.~\ref{fig:teaser}.

In Table~\ref{tab:design_consideration}, the initial configuration, ``Slice Table \& Slice Detail View,'' is capable of presenting slice performance indicators and instance attributions. However, the numerical matrices of performance indicators are not sufficient in guiding users to uncover hidden model issues, such as spurious correlations that can happen regardless of high or low performance. A subsequent iteration included a scatter plot showing the attribution representation space. This configuration enables the system's instance performance indicator capability but still misses the slice attribution overview. To the best of our knowledge, there exist few approaches that visually elucidate potential model spuriousness in scale, where our third configuration fulfills this gap --- ``Slice Table \& Slice Detail View \& \mosaicname{}, '' which comprehensively satisfies all the essential system capabilities and fulfills our design requirements.

\mosaicname{}, alongside the Slice Table and Slice Detail View, enriched the system's ability to provide granular insight into both slice and instance-level attributions and performance indicators (\ref{req:interpretability}, \ref{req:summary}). The coordinated Slice Table and \mosaicname{} can guide users to pinpoint hidden issues, and the Slice Detail View enables users to further verify their insights (\ref{req:actionable}). This robust configuration underscores our meticulous design process in assuring that \systemname{} is not only capable of furnishing critical insights but also does so in an intuitive and user-friendly manner.

The tabulated design considerations and the ensuing choice of system component configuration aim at maximizing the system's utility and user-centric functionality. With our finalized configuration (the last row of Table~\ref{tab:design_consideration}), users are empowered with sufficient support for problem detection, interpretation, and mitigation in the model validation.

\subsection{Slice Error Mitigation for Model Enhancement} \label{subsec:error_mitigation}
% \textbf{Model Retraining.} 
One effective approach to mitigating spurious correlations in ML models is through model re-training, which can improve the model's robustness without changing its architecture. We leverage the Core Risk Minimization (CoRM) method introduced in~\cite{singla2022core}. CoRM corrupts non-core image regions with random Gaussian noise and retrains the model using the noise-corrupted data, which has been shown to be effective in mitigating a model's reliance on spurious features.

We first export slices with high Spuriousness probabilities, indicating undesired correlations. For the images in these slices, the model attribution masks (e.g., GradCAM masks) highlight spurious regions and we utilize such masks to add random Gaussian noise to spurious regions. For a single image, this process can be represented by $\mathbf{x'} = \mathbf{x} + \mathbf{m} \bigodot \mathbf{z}$, where $\mathbf{x}$ is the input image, $\mathbf{m}$ is the attribution mask, and $\mathbf{z}$ is the generated Gaussian noise matrix. All these three variables are of the same size as the input image, and $\bigodot$ denotes the Hadammard product. Fig.~\ref{fig:spurious_CoRM} shows some examples of this operation, with exaggerated noise for presentation purposes. After replacing the original data with these noisy-corrupted slices, we retrain the model and evaluate whether spurious correlation has been reduced (\ref{req:actionable}). In Sec.~\ref{sec:evaluation}, we explain the evaluation metrics we use to quantify the effectiveness of \systemname{} in mitigating spurious correlations.

\section{Case Studies}
\label{sec:usecases}
\begin{figure*}
    \centering
    \includegraphics[width=0.85\linewidth]{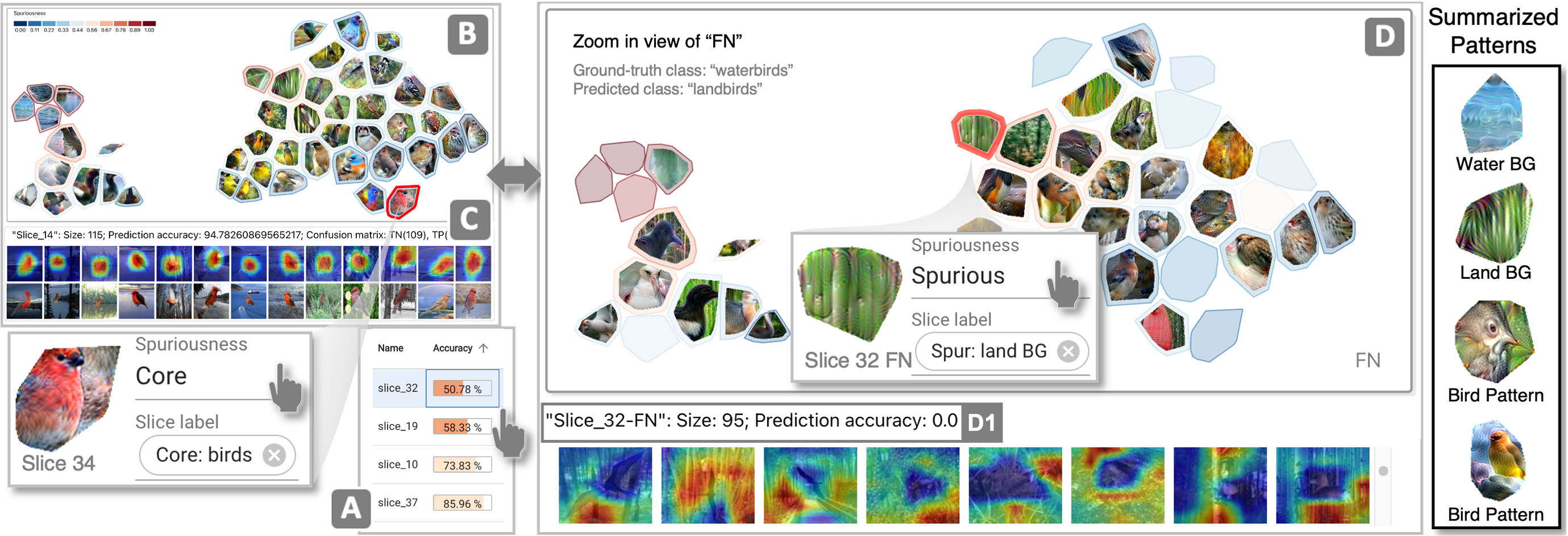}
    \caption{
    \systemname{} applied to the validation of a bird category classification model trained on the Waterbirds dataset. The user finds the correct model behavior (bird patterns) corresponding to a well-performed \textit{Slice 34} and annotates it as ``core: birds'' \mybox{\textcolor{white}{B}} and \mybox{\textcolor{white}{C}}. By switching the \mosaicname{} to the confusion matrix view \mybox{\textcolor{white}{D}}  and investigating the underperformed slices with accuracy sorting \mybox{\textcolor{white}{A}}, the user identifies a problematic \textit{Slice 32} that has high false negatives \mybox{\textcolor{white}{D1}}, which turns out to use spurious feature of land backgrounds (BG) to predict landbirds.
    }
    \label{fig:use_case_2}
\end{figure*}
\begin{figure}[!bth]
    \centering
    \includegraphics[width=\linewidth]{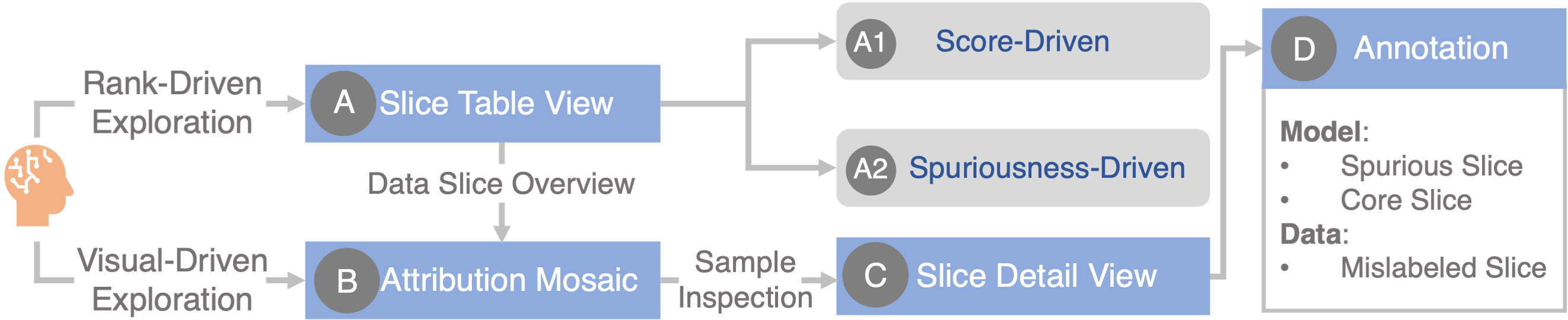}
    \caption{Usage patterns for \systemname{} derived from case studies. The analysis can start with \myboxx{\textcolor{white}{A}} a rank-driven exploration (ranking the slices by \mylongboxT{\textcolor{white}{{\tiny A1}}} score (e.g., model accuracy) or \mylongboxT{\textcolor{white}{{\tiny A2}}} spuriousness), or \myboxx{\textcolor{white}{B}} a visual-driven exploration. After inspecting slice summary in \mosaicname{} (\myboxx{\textcolor{white}{B}}), they can inspect individual samples in the Slice Detail View (\myboxx{\textcolor{white}{C}}). Finally, they can annotate their insights~(\myboxx{\textcolor{white}{D}}).}
    \label{fig:usage_workflow}
\end{figure}

In this section, we present two case studies with publicly available vision datasets to benchmark and evaluate the capabilities of \systemname{}. The primary objective of these case studies is to demonstrate how \systemname{} empowers ML experts and practitioners to detect, evaluate, and interpret potential issues in vision models. Fig.~\ref{fig:usage_workflow} illustrates common usage patterns of our system: users can begin the analysis with either a rank-driven evaluation \myboxxA{\textcolor{white}{A}} or a visual-driven evaluation~\myboxxA{\textcolor{white}{B}}. Once a slice of interest is identified, users can delve deeper into individual samples in the Slice Detail View~\myboxxA{\textcolor{white}{C}}. Finally, the user can annotate the data slices and continue their analysis~\myboxxA{\textcolor{white}{D}}.  

\subsection{Hair Color Classifier Validation - Finding Edge Cases}\label{subsec:use_case_1} 

\noindent\textbf{$\bullet$ Overview.} This case study involves the Large-scale CelebFaces Attributes (CelebA) dataset~\cite{liu2015faceattributes} with $202,599$ number of face images. The label of each image is one of \{\textit{not gray hair}, \textit{gray hair}\}, refer to labels $\{0, 1\}$, respectively. With the train, validation, and test splits of $(8:1:1)$, we adopt transfer learning to train a ResNet50~\cite{he2016deep} binary image classifier. After iteratively fine-tuning hyper-parameters, we obtain a model with $98.03\%$ classification accuracy. The ML experts would then explain and troubleshoot this hair color classifier with \systemname{} based on these settings: \textit{n\_neighbors} $=5$, \textit{min\_dist} $=0.01$, and \textit{n\_components} $=2$ for the UMAP algorithm, and \textit{n\_clusters} $=50$ for K-Means. 

\begin{figure}[!thb]
  \centering
  \includegraphics[width=.99\linewidth]{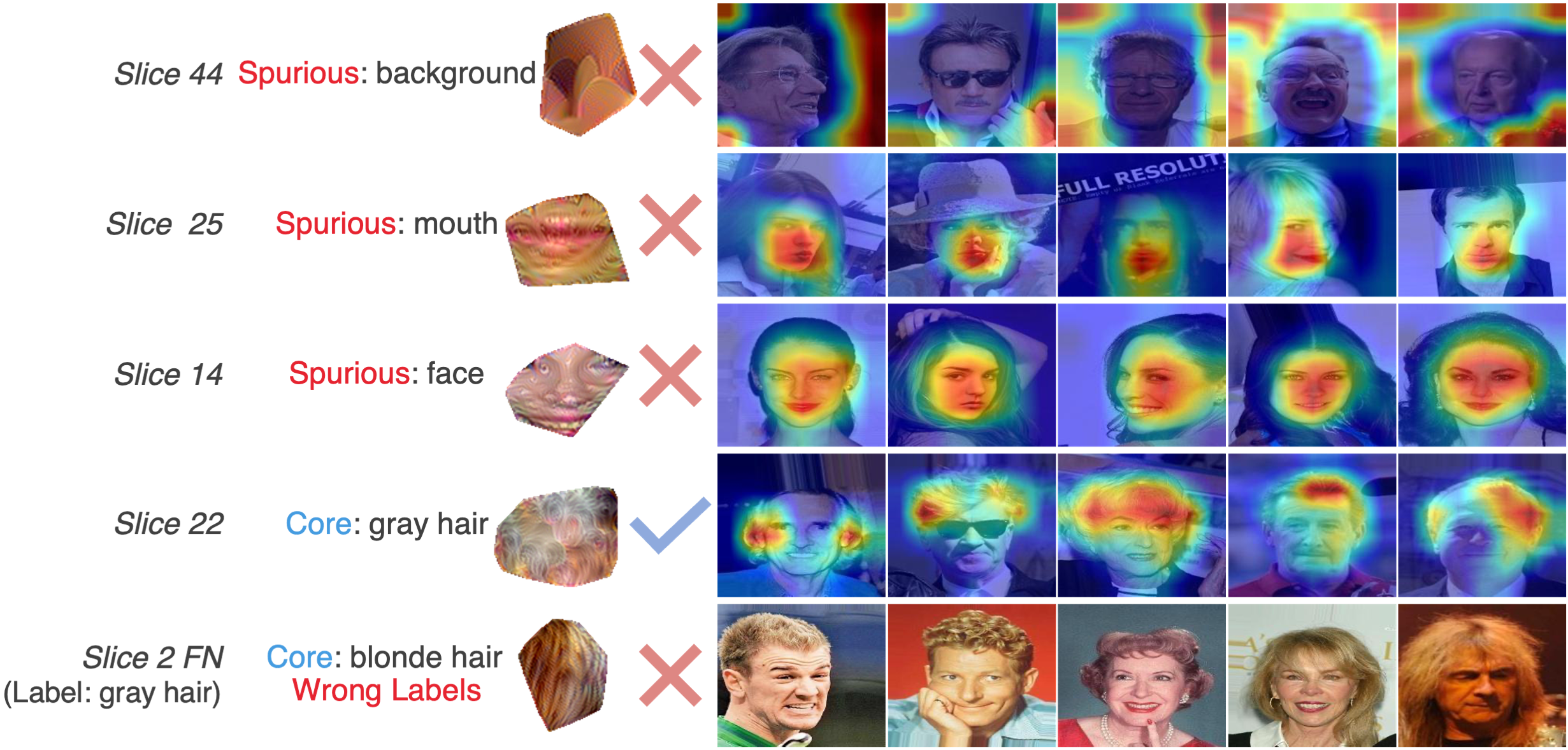}
  \caption{Examples of findings when ML experts validate a hair color classifier, where experts discover slices corresponding to the model's core/spurious correlations and wrong labels.
  }
  \label{fig:case_1_insight}
\end{figure}

\noindent\textbf{$\bullet$ Does the model behave correctly on well-performing slices?} One basic expected behavior for a hair color classification model is to catch hair features. At first glance of the \mosaicname{} (Fig.~\ref{fig:teaser}~\mybox{\textcolor{white}{C}}), ML experts notice \textit{slice\_22} and \textit{slice\_5} on the left, which lie separately with others on the \mosaicname{}.
Their mosaic visualizations indicate gray hair patterns, which means the model uses the correct features, i.e., core features. By verifying the corresponding attribution heatmaps (Fig.~\ref{fig:teaser}~\mybox{\textcolor{white}{D}}), they confirm the correctness of this insight and annotate both slices as \textit{core feature} with description \textit{Core: gray hair}. Upon experts saving the annotation, \systemname{} automatically propagates the annotation and provides a Spuriousness probability of each slice. With this guidance, the experts observe many slices lying on the right of \mosaicname{} have higher Spuriousness probabilities (Fig.~\ref{fig:teaser}~\mybox{\textcolor{white}{C}}~\mybox{\textcolor{white}{E}}). Their mosaic visualizations do not show any hair patterns, leading to valid doubts that the model does not behave correctly on such slices. Besides, they also notice that the model has correct predictions on these slices (with $100\%$ prediction accuracy). This makes them worry that the model is largely biased by spurious features. Through investigation, they find the model mistakenly utilizes mouth and eyes to predict hair color for those top-performed slices, such as \textit{slice\_25}, \textit{slice\_14}, and \textit{slice\_2} (Fig.~\ref{fig:case_1_insight}).

\noindent\textbf{$\bullet$ What underlying factors contribute to unexpected behaviors?} By sorting the Slice Table by descending order of the propagated \textit{Spuriousness}, ML experts notice more slices with high Spuriousness, such as \textit{slice\_35}, and are interested in understanding why such unexpected model behaviors happen. They switch \mosaicname{} into the confusion matrix form (Fig.~\ref{fig:teaser}~\mybox{\textcolor{white}{E}}) and click on the name of ``slice\_35'' to highlight this slice on the four sub-views and inspect explanations. They notice that images that lie in the ``FN'' group of this slice have wrong labels --- they should be labeled as ``not gray hair'' rather than ``gray hair''. They mark this issue as wrong labels and investigate its neighborhood slices on \mosaicname{}, finding \textit{slice\_2\_FN} also features wrong labels.

\noindent\textbf{$\bullet$ Why slices underperform?} After fully exploring those well-performed but biased slices, the experts wonder whether the underperformed slices are indeed issue-free or what contributes to their low performance. By sorting the Slice Table (Fig.~\ref{fig:teaser}~\mybox{\textcolor{white}{B}}) by ascending order of accuracy, they notice the model only achieves $72.41\%$ accuracy on \textit{slice\_44} and find that the slice's visualization only shows colorful patterns without any recognizable content. After checking the attribution heatmaps, they find that the model mistakenly uses image backgrounds to make hair color predictions (Fig.~\ref{fig:case_1_insight}). This spurious correlation issue stands out, and they annotate this slice as ``spurious'' with the description ``Spurious: backgrounds''. Similar issues also occur in its neighborhood slices via~\mosaicname{}, and \systemname{} automatically propagates them with higher Spuriousness.

\noindent\textbf{$\bullet$ Insights summary.} Fig.~\ref{fig:case_1_insight} presents examples of the detected issues with the current model, which include wrong attributions in underperforming slices, model biases in well-performing slices, and noisy label issues. Through the aforementioned procedure in this case study, we demonstrate how \systemname{} supports users to uncover and interpret potential model issues with visual summaries and other guidance. Based on annotations from ML experts, we employ the CoRM framework to mitigate the detected errors, which will be evaluated in Sec.~\ref{sec:evaluation}.

\subsection{Bird Category Classifier Validation - Detecting Bias}\label{subsec:use_case_2}
% for AI Fairness

\noindent\textbf{$\bullet$ Overview.} To study whether \systemname{} can help ML experts and practitioners find the potential biases and discrimination of models, we design this case study with a biased dataset called Waterbirds~\cite{sagawa2019distributionally}, which is constructed by cropping out birds from images in the Caltech-UCSD Birds-200-2011 (CUB) dataset~\cite{wah2011caltech} and transferring them onto backgrounds from the Places dataset~\cite{zhou2017places}. For each image, the label belongs to one of \{\textit{waterbird}, \textit{landbird}\}, and the image background belongs to one of \{\textit{water background}, \textit{land background}\}. The training set is skewed by placing $95\%$ of waterbirds (landbirds) against a water (land) background and the remaining $5\%$ against a land (water) background. Following the same data splitting as~\cite{sagawa2019distributionally}, the train, validation, and test sets include $4795$, $1199$, and $5794$ images, respectively. After training and fine-tuning hyper-parameters, the waterbirds/landbirds classification model achieves $85.74\%$ classification accuracy. In the data slice finding, we set \textit{n\_neighbors} $=20$, \textit{min\_dist} $=0.05$, and \textit{n\_components} $=2$ for the UMAP algorithm, and \textit{n\_clusters} $=46$ for K-Means.

While in this study, ML experts are aware that the model is likely biased by backgrounds --- using water (land) background to classify waterbirds (landbirds). However, such priori knowledge is hard to establish in real-world applications because of the scarcity of additional well-labeled metadata. And hence the experts assume such information is unknown and want to validate whether \systemname{} can make the potential model biases stand out by only utilizing the original images and the trained model.

\begin{figure}[!t]
  \centering
  \includegraphics[width=.91\linewidth]{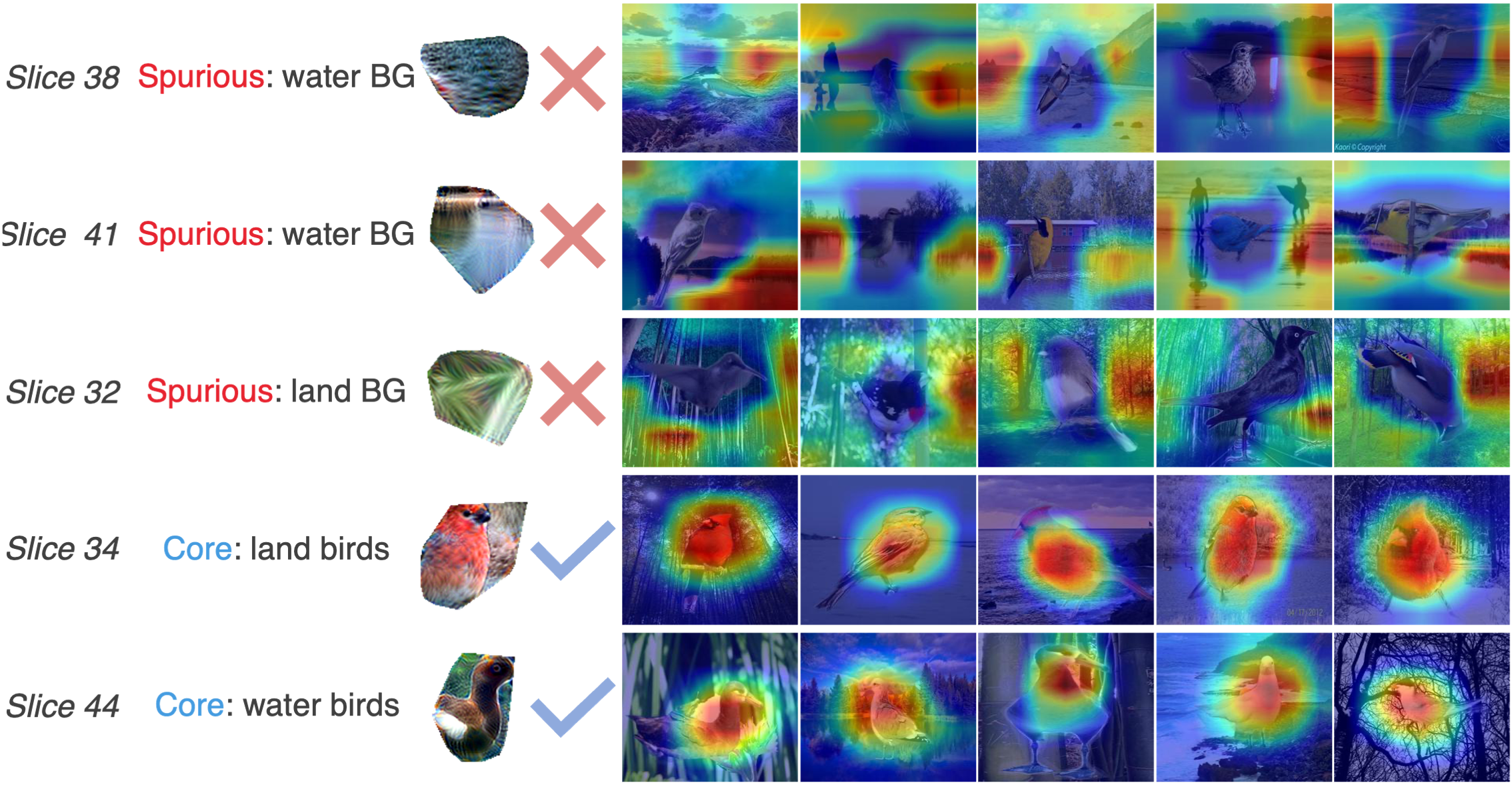}
  \caption{Examples of findings when ML experts validate a bird category classifier, where experts identify problematic slices where the model uses backgrounds (BG) to classify birds.}
  \label{fig:case_2_insight}
\end{figure}

\noindent\textbf{$\bullet$ Does the model exhibit bias?} To answer this key question, experts start by investigating the underperformed slices. From the Slice Table (Fig.~\ref{fig:use_case_2}~\mybox{\textcolor{white}{A}}), they sort slices by ascending order of accuracy and select the worst-performed \textit{slice\_38}. The coordinated information provided by \mosaicname{} and model attribution heatmaps highlights a spurious correlation problem---the model uses water backgrounds to classify birds (Fig.~\ref{fig:case_2_insight}).
The experts annotate this slice as ``spurious'', and \systemname{} automatically propagates this annotation. They verify the propagation correctness on neighborhood slices and annotate \textit{slice\_41} as ``spurious'' with ``water backgrounds'' (Fig.~\ref{fig:case_2_insight}). Moreover, the experts identify an underperformed \textit{slice\_32} that is not clustered with the current ones and is given a high Spuriousness possibility. They investigate it and verify that the model utilizes ``Spurious feature: land backgrounds'', to predict bird classes. Through the \mosaicname{} in the confusion matrix form (Fig.~\ref{fig:use_case_2}~\mybox{\textcolor{white}{D}}), they find such spurious correlations result in many false negatives (Fig.~\ref{fig:use_case_2}~\mybox{\textcolor{white}{D1}}), where the model uses land backgrounds to mistakenly predict many ``waterbirds'' as ``landbirds''.

\noindent\textbf{$\bullet$ Is the detected bias prevalent across all slices? Why or why not?} ML experts are interested in determining whether the detected bias is pervasive throughout the dataset.
By analyzing slices that are distant from the annotated ones in the \mosaicname{} and are assigned with low Spuriousness possibilities, they discovered that the farthest neighbors, namely \textit{slice\_34} and \textit{slice\_5}, correspond to core features. This suggests that the model can correctly capture bird regions in these slices (Fig.~\ref{fig:use_case_2}~\mybox{\textcolor{white}{B}}~\mybox{\textcolor{white}{C}}), which raises a follow-up ``why'' question. To understand in what circumstances when the model fails. They browse the original images from \textit{slice\_32} (spurious feature) and \textit{slice\_34} (core feature), respectively, as shown in Fig.~\ref{fig:use_case_2}~\mybox{\textcolor{white}{D}} and~\mybox{\textcolor{white}{C}}. They find \textit{slice\_32} has very similar land backgrounds and very different birds, while on the other hand, the birds' appearance of \textit{slice\_34} is very consistent. This finding explains why this biased model can successfully capture the core features from \textit{slice\_34} but fails at \textit{slice\_32}---greater similarities in the representation space indicate stronger features. In other words, core features in \textit{slice\_34} are strong enough to support the model's robustness against bias. Such insights are helpful in improving model robustness and have been further studied by ML experts~\cite{zhang2022correct}.

\noindent\textbf{$\bullet$ Insights summary.} A summary of insights obtained from this case study is provided in Fig.~\ref{fig:case_2_insight}, where ML experts validate the existence of model biases and extract slices corresponding to different biases. In the following Sec.~\ref{sec:evaluation}, we evaluate whether \systemname{} can mitigate model errors by incorporating human feedback.

\begin{figure*}
  \centering
  \includegraphics[width=.9\linewidth]{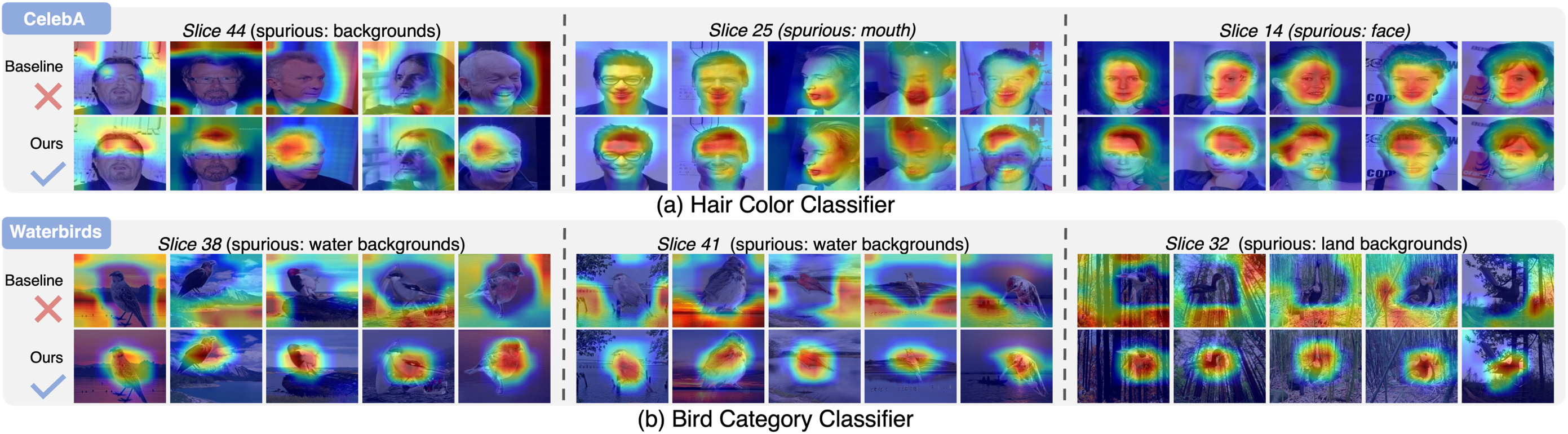}
  \caption{Qualitative evaluation with GradCAM attribution heatmaps. We visually compare the original and the \systemname{}-improved models for hair color and bird category classification, respectively. In each sub-figure, the first row refers to the original model (baseline), and the second row refers to the improved model (ours). We can observe \systemname{} suppresses the spurious correlations by using the correct features for predictions.}
  \label{fig:eval}
\end{figure*}

\section{Evaluation}
\label{sec:evaluation}
% In this section, we conduct both quantitative and qualitative evaluations to validate whether \systemname{} can indeed leverage human insights to enhance vision models' performance while reducing their reliance on spurious features.

\subsection{Quantitative Evaluation}\label{subsec:quantified_eval}

The quantitative evaluation involves four matrices introduced in~\cite{singla2022core}, including clean accuracy, core accuracy, spurious accuracy, and relative core sensitivity (RCS). We revisit their definitions as follows:

% \vspace{.5mm}
\noindent\textbf{$\bullet$ Clean Accuracy.}
The original model accuracy, where larger values are indicative of better overall accuracy.

% \vspace{.5mm}
\noindent\textbf{$\bullet$ Core Accuracy $acc^{(C)}$.}
The model accuracy when spurious regions are occupied with Gaussian noise, where larger values are indicative of the model's more reliance on core regions.

% \vspace{.5mm}
\noindent\textbf{$\bullet$ Spurious Accuracy $acc^{(S)}$.}
The model accuracy when core regions are added with Gaussian noise, where larger values are indicative of the model's more reliance on spurious regions.

% \vspace{.5mm}
\noindent\textbf{$\bullet$ RCS.}
This metric quantifies the model's reliance on core features while controlling for general noise robustness. It is defined as the ratio of the absolute gap between core and spurious accuracy, and the total possible gap for any model between core and spurious accuracy. Represented as $RCS=\frac{acc^{(C)} - acc^{(S)}}{2 \times min(\alpha, 1-\alpha)}$, where $\alpha = \frac{acc^{(C)} + acc^{(S)}}{2}$. RCS ranges from $0$ to $1$, a higher value indicative of better model performance.
% (A comprehensive proof can be found in~\cite{singla2022core}.)

\begin{table}[]
\centering
\caption{Quantitative evaluation of the performance of the hair color classification models trained on the original and \systemname{}(\emph{AS})-improved CelebA dataset.}
\label{tab:celebA_eval_table}
\resizebox{\columnwidth}{!}{%
\begin{tabular}{ccccc}
\rowcolor[HTML]{5F97D2} 
{\color[HTML]{FFFFFF} \textbf{Training Procedure}}                         & {\color[HTML]{FFFFFF} \textbf{Clean Acc ($\uparrow$)}} & {\color[HTML]{FFFFFF} \textbf{Core Acc ($\uparrow$)} } & {\color[HTML]{FFFFFF} \textbf{Spurious Acc ($\downarrow$)}} & {\color[HTML]{FFFFFF} \textbf{RCS ($\uparrow$)}} \\
\rowcolor[HTML]{E6F0FF} 
    Baseline                                                                        & 98.02688                                  & 97.21170                                  & 97.61917                                     & 0.067720                            \\
\rowcolor[HTML]{F8FAFE} 
\emph{AS} (Annotation) & 98.02688                                  & 98.02185                                 & 96.92958                                     & 0.216351                            \\
\rowcolor[HTML]{E6F0FF} 
\emph{AS} (Propagation) & \textbf{98.08225}                         & \textbf{98.16278}                        & \textbf{96.01852}                            & \textbf{0.368512}                  
\end{tabular}%
}
\end{table}

% \vspace{1mm}
In the two case studies discussed in Sec.~\ref{sec:usecases}, ML experts annotated five spurious slices in the CelebA dataset, \{\textit{slice\_25, slice\_14, slice\_2, slice\_44, slice\_26}\}, and six spurious slices in the Waterbirds dataset, \{\textit{slice\_38, slice\_41, slice\_32, slice\_29, slice\_4, slice\_10}\}, respectively. For each case study, \systemname{} automatically ran the Label Propagation algorithm and exported both the users' annotation records and the propagated Spuriousness for further investigation.

To thoroughly evaluate our introduced method, we validate three models for each case. The models marked as ``baseline'' are the original trained models obtained at the beginning of each case study. The models marked as ``\systemname{}'' are re-trained using the CoRM method after adding noise to ``spurious'' slices according to results exported from \systemname{}. In particular, ``Annotation'' indicates that only user-annotated spurious slices were corrupted with noise, while ``Propagation'' indicates that propagated Spuriousness is used to identify spurious slices to be added with noise. 

Table~\ref{tab:celebA_eval_table} and Table~\ref{tab:waterbird_eval_table} present the quantitative evaluation results for the two case studies, respectively. Our presented \systemname{} can significantly improve the vision model's overall performance with reduced spurious correlations. Furthermore, our Label Propagation largely reduces human effort by automating the annotation process, and achieves the best performance in this quantified evaluation. Overall, our results demonstrate that \systemname{} is an effective approach for mitigating spurious correlations in machine learning models, and the Label Propagation algorithm is a valuable tool for automating the annotation process.

% \vspace{-10pt}
\subsection{Qualitative Evaluation} \label{subsec:qualified_eval}

Results in Sec.~\ref{subsec:quantified_eval} highlight the best performance of \systemname{} equipped with Label Propagation in mitigating spurious correlations. For further evaluation, we visually compare models' attributions via GradCAM in Fig.~\ref{fig:eval}. Refer to Sec.~\ref{subsec:use_case_1}, the hair color classifier originally has spurious correlations dominant in \{\textit{slice\_44, slice\_25, and slice\_14}\}, where the model uses image backgrounds, mouth, or eyes to predict hair color, respectively. With the help of \systemname{}, the model's misattribution is successfully fixed, directing focus towards the correct hair regions (refer to Fig.~\ref{fig:eval}(a)). As for the bird category classification model, it originally focuses on spurious features, water/land backgrounds, to decide whether there are water/land birds on the input image \{\textit{slice\_38, slice\_41, and slice\_32}\} (refer to Sec.~\ref{subsec:use_case_1}). In Fig.~\ref{fig:eval}(b), we can observe that \systemname{} mitigates these issues by helping the model to focus on the core bird features.

\begin{table}[!t]
\centering
\caption{Quantitative evaluation of the performance of the bird category classification models trained on the original and \systemname{}(\emph{AS})-improved Waterbirds dataset.}
\label{tab:waterbird_eval_table}
\resizebox{\columnwidth}{!}{%
\begin{tabular}{ccccc}
\rowcolor[HTML]{5F97D2} 
{\color[HTML]{FFFFFF} \textbf{Training Procedure}}                         & {\color[HTML]{FFFFFF} \textbf{Clean Acc ($\uparrow$)}} & {\color[HTML]{FFFFFF} \textbf{Core Acc ($\uparrow$)} } & {\color[HTML]{FFFFFF} \textbf{Spurious Acc ($\downarrow$)}} & {\color[HTML]{FFFFFF} \textbf{RCS ($\uparrow$)}} \\
\rowcolor[HTML]{E6F0FF} 
Baseline                                                                        & 85.73812                                  & 82.98582                                  & 82.82901                                     & 0.004878                            \\
\rowcolor[HTML]{F8FAFE} 
\begin{tabular}[c]{@{}c@{}}\emph{AS} (Annotation)\end{tabular} & 89.57465                                  & 86.40534                                 & 79.81651                                     & 0.195062                            \\
\rowcolor[HTML]{E6F0FF} 
\begin{tabular}[c]{@{}c@{}}\emph{AS} (Propagation)\end{tabular} & \textbf{90.40867}                         & \textbf{87.40617}                        & \textbf{77.89825}                            & \textbf{0.274038}                  
\end{tabular}%
}
\end{table}

% \vspace{-10pt}
\subsection{Experts Feedback}
We conduct two case studies (see Sec.~\ref{subsec:use_case_1} and Sec.~\ref{subsec:use_case_2}) with ten ML experts to evaluate \systemname{}. None of them are co-authors of this paper, and they had not previously seen \systemname{}. Five are male and five are female. Six of the experts have over five years of experience in ML, two have between three to five years of experience, and two have between one to three years of experience. All experts need to conduct model validation in their research and have experience working with vision models. Our findings are drawn from their comments during the study, as they follow the ``think-aloud'' protocol, and additional feedback is gathered through exit discussion and questionnaire.

\begin{figure}
    \centering
    \includegraphics[width=1\linewidth]{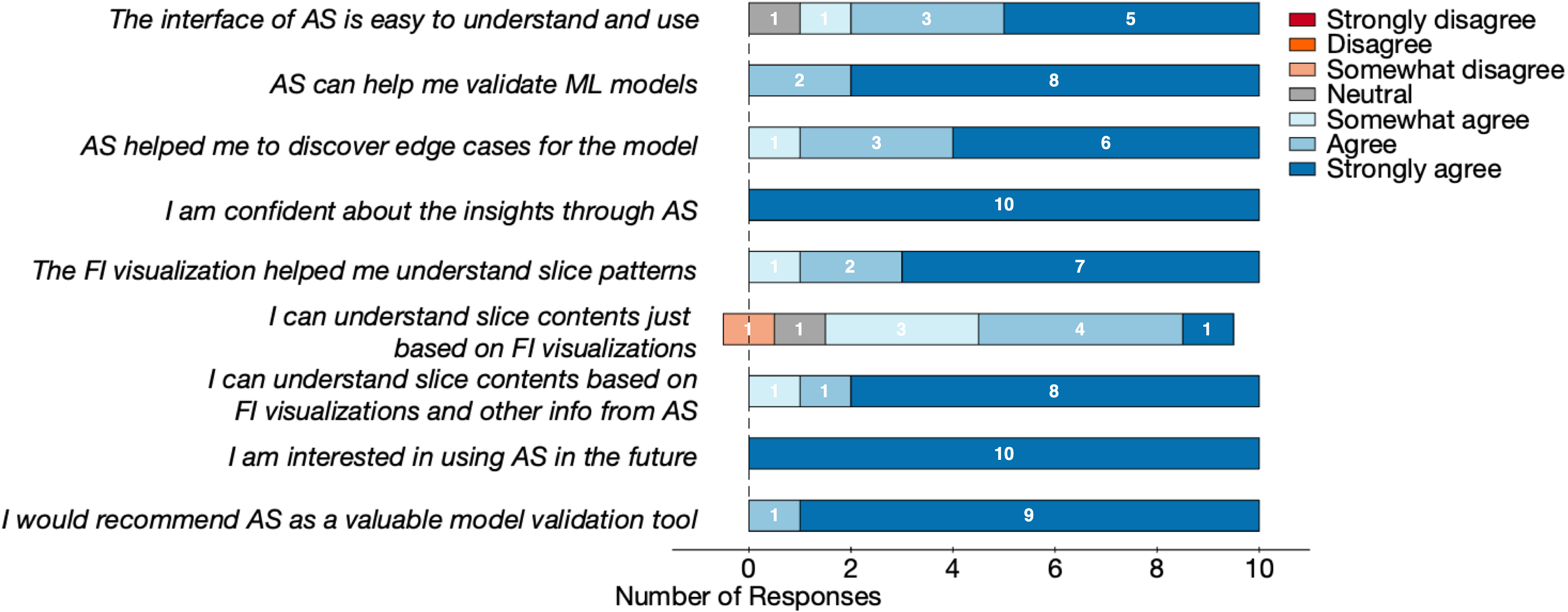}
    \caption{Expert perception of the system, according to eight Likert-Type Questions using a 7-point scale. (\emph{AS} = \systemname{}.)
    }
    \label{fig:likert}
\end{figure}

% \vspace{.5mm}
\textbf{Summary of Likert-Type Questions.}\hspace{3mm}
The exit questionnaire included nine Likert-type questions on a seven-point scale, aiming to evaluate \systemname{} from various aspects. The feedback, as shown in Fig.~\ref{fig:likert}, reflects a positive overall impression. Notably, in the usability evaluation, five experts strongly agreed and three agreed that our system is user-friendly and understandable. In terms of effectiveness, eight experts strongly agreed that our system aids in validating models. All experts expressed strong confidence in the insights provided. Considering that Feature Inversion (FI) results are synthesized images that might initially be hard to grasp, we included questions to assess this design aspect. Seven experts strongly agreed, and two agreed that FI is helpful for understanding slice patterns. While two experts expressed neutral or somewhat disagreeing opinions about understanding slice contents based solely on FI, eight strongly agreed that \systemname{}'s coordination of FI and other views supports a comprehensive understanding of slice contents. Lastly, all experts expressed interest in using our system in the future and would recommend it as a valuable tool for model validation.

% \vspace{.5mm}
\textbf{Interpretable Model Validation.}\hspace{3mm}
All experts acknowledged that \systemname{} ``\textit{effectively helps ML model validation}''. They remarked that the complimentary visual summaries and attribution heatmaps help them ``\textit{easily understand slice issues}''. ``\textit{It's often hard to figure out what's going wrong with a model}'', one expert said, ``\textit{This system guides me to find where I should look and explains the issues intuitively.}'' Another expert noted, ``\textit{The ability to browse what's happening in different slices at once greatly helped me understand the model.}'' They commented that they are confident about their insights because our system ``\textit{highlights slice errors and provides supportive evidence}''. Three of them are eager to see what issues \systemname{} could uncover for the models they are currently using.

% \vspace{.5mm}
\textbf{\mosaicname{}.}\hspace{3mm}
All participants showed particular interest in the \mosaicname{} and asked about how it was designed in the interview. Seven experts were impressed by its ability to summarize model attributions across data subgroups. Two experts noted that ``\textit{it highlights model issues intuitively.}'' Another described it as ``\textit{impressive and novel},'' emphasizing that it ``\textit{highlights and explains the model's failures}.'' Another expert commented, ``\textit{This view provides a new angle for me to identify and understand the data/model issues}.'' All experts quickly adapted to using Spuriousness propagation and utilized the Spuriousness matrix to speed up their annotation. Specifically, five experts found that the Spuriousness propagation ``\textit{surprisingly helpful},'' and six remarked that it ``\textit{really saved my effort}''. 

% One expert who is not familiar with XAI techniques mentioned ``I didn't understand the patterns at the first glance, but by checking out the neighborhood mosaics, it becomes clear and is very helpful for me to validate the problems.''

% \section{Discussion and Future Work}
% \label{sec:discussion}
% \input{Sections/6-Discussion}

\section{Conclusion}
\label{sec:conclusion}
In this work, we present \systemname{}, a novel human-in-the-loop model validation system empowered by metadata-free data slice finding and XAI techniques, which can support effective detection, explanation, and mitigation of potential errors and biases in vision models. 
\systemname{} provides explainable data slices that reveal critical model problems, such as spurious correlations and mislabeled data. The effectiveness of this work is validated by the performance improvement and bias mitigation through both quantitative and qualitative evaluations. We hope this work can inspire future research in human-AI teaming to improve AI trustworthiness and accountability.

\section*{Acknowledgments}
This research is supported in part by the National Science Foundation via 
grant No. IIS-2427770, the National Institute of Health under grant no. P41 EB032840, and the University of California (UC) Multicampus Research Programs and Initiatives (MRPI) grant and the UC Climate Action Initiative grant.

%%%%%%%%%%%%%%%%%%%%%%%%%%%%%%%%%%%%%%%%% \uparrow comment this out for appendix %%%%%%%%%%%%%%%%%%%%%%%%%%%%%%%%%%%%%%%%%

% % %%%%%%%%%%%%%%%%%%%%%%%%%%%%%%%%%%%%%%% \downarrow comment this out for main paper %%%%%%%%%%%%%%%%%%%%%%%%%%%%%%%%%%%%%%%
% \newpage
{\appendices
% \setcounter{page}{1}
% \twocolumn[
%         \centering
%         \Large
%         \textbf{\thetitle}\\
%         \vspace{0.5em}Appendix\\
%         \vspace{1.0em}
% ]

\section*{Appendix}

\setcounter{section}{0}%
\setcounter{subsection}{0}%
\setcounter{subsubsection}{0}%
\setcounter{equation}{0}
\renewcommand\theequation{A\arabic{equation}} 
\setcounter{table}{0}  
\setcounter{figure}{0}
\renewcommand{\thetable}{A\arabic{table}}
\renewcommand{\thefigure}{A\arabic{figure}}
\renewcommand{\thealgorithm}{A\arabic{algorithm}}
\newcommand\appendixsection[1]{%
    \stepcounter{section}
    \renewcommand{\thesection}{A-\Roman{section}}
    \section*{\thesection.~#1}
    \hypertarget{A-\Roman{section}}{}
}

\noindent This appendix is organized as follows:
\begin{itemize}
    \item Sec.~\hyperlink{A-I}{A-I} provides equations for calculating neighbor consistency of data features and model attributions, which are used to compare two feature representation spaces.
    \item Sec.~\hyperlink{A-II}{A-II} provides equations for calculating slice attribution consistency.
    \item Sec.~\hyperlink{A-III}{A-III} explains our interactive visual layer adjustment in the design of \mosaicname{}.
    \item Sec.~\hyperlink{A-IV}{A-IV} details the Label Spreading Algorithm to complement the main paper.
    \item Sec.~\hyperlink{A-V}{A-V} provides additional feedback from our experts.
    \item Sec.~\hyperlink{A-VI}{A-VI} provides the discussion and future work.
    
\end{itemize}

\appendixsection{Representation Space Comparison}
\label{appendix_sec:space_comparison}
In this section, we provide equations for calculating the local consistency of data features and model attributions, to complement the comparative study between the \textbf{feature representation space} (derived from the original feature vectors) and the \textbf{attribution representation space} (derived from the attribution-weighted feature vectors) as discussed in Sec. III-D(2) and Fig. 4 of the main paper.

Given a specific data sample \(i\) and its neighbor sample \(j\) on a representation space, we denote the original feature vector produced by model's convolutional modules as \(F_*\), and its GradCAM attribution mask as \(M_*\). The data feature similarity between \(i\) and \(j\) is computed by:
\begin{equation}
    D_{sim}(F_i, F_j) = \frac{\langle F_i, F_j \rangle}{\| F_i \| \cdot \| F_j \|},
\end{equation}
which is the cosine similarity between \(F_i\) and \(F_j\), with a higher value indicative of higher similarity.
The model attribution similarity between \(i\) and \(j\) is computed by:
\begin{equation}
    A_{sim}(M_i, M_j) = \frac{1}{W(M_i, M_j)},
\end{equation}
where \(W(M_i, M_j)\) represents the Wasserstein distance (also referred to as Earth Mover's Distance) between the GradCAM attribution masks \(M_i\) and \(M_j\).

Then, for a specific sample \(i\), the local consistency of data features is computed by the average \(D_{sim}(F_i, F_*)\), and the local consistency of model attributions is computed by the average \(A_{sim}(M_i, M_*)\), with higher values indicative of better local consistency, where \(*\) denotes the top 10 nearest neighbors of \(i\) on the corresponding representation space.

\appendixsection{Slice Attribution Consistency}
\label{appendix_sec:slice_consistency}
To complement the data slice identification in Sec. III-D(2) of the main paper, we provide equations for calculating the attribution consistency of a data slice. Given a data slice \(S_i\) containing \(N\) attribution-weighted feature vectors, each denoted by \(F_W\), the slice centroid is computed as:
\begin{equation}
    S_{i}^{c} = \frac{1}{N}\sum_{j=1}^{N}{F_W^j}, F_W^j \in S_i,
\end{equation}
The attribution consistency of this slice is then calculated as:
\begin{equation}
    C_{S_i} = \frac{\sum_{j=1}^{N}{sim(F_W^j, S_{i}^{c})}}{N}, F_W^j \in S_i,
\end{equation}
where \(sim(\cdot, \cdot)\) denotes cosine similarity.

\appendixsection{Interactive Visual Layer Adjustment}

In our \mosaicname{} (refer to Sec. III-E of the main paper), we provide interactive visual layer adjustment to further support user exploration of our mosaic view. When hovering over a specific data slice, the corresponding slice mosaic would automatically expand and display on the top visual layer to avoid being covered by other mosaic panels. This automatic visual layer adjustment ensures clear visibility of all data slices in the mosaic view to facilitate user exploration.

\appendixsection{Label Spreading Algorithm}
\label{appendix_sec:label_spreading}

We detail the description of the Label Spreading Algorithm from~\cite{zhou2003learning} to complement Sec. III-E of the main paper.

The Label Spreading algorithm is a semi-supervised approach that propagates label information from a few annotated data points to unannotated ones by leveraging the similarity structure of the data. It assumes that similar data points should have similar labels and iteratively updates labels to enforce both local and global consistency. The key steps include:

\noindent\textbf{1. Graph Construction.}
   A graph is constructed where each data point is represented as a node, and the edge weights represent pairwise similarities between data points. The similarity between \(x_i\) and \(x_j\) can be defined based on different requirements, such as using a Gaussian kernel:
   \[
   w_{ij} = \exp\left(-\frac{\|x_i - x_j\|^2}{2\sigma^2}\right),
   \]
   where \(\|x_i - x_j\|\) is the Euclidean distance, and \(\sigma\) is the scaling parameter that controls the kernel width. In our case, we fit the graph to our attribution representation space and \(x_*\) is the 2D coordinates of a sample on the space.
   
\noindent\textbf{2. Graph Normalization.}
   The weight matrix \(W = [w_{ij}]\) is normalized to compute the transition matrix \(T = D^{-1}W\), and \(D\) is the diagonal degree matrix with \(D_{ii} = \sum_j w_{ij}\).

\noindent\textbf{3. Label Initialization.}
   A label matrix \(Y\) is initialized, where initial labels are binary values \(\{0, 1\}\) for annotated points and \(-1\) for unannotated points.

\noindent\textbf{4. Iterative Propagation.}
   Labels are updated iteratively using:
   \[
   Y^{(t+1)} = \alpha T Y^{(t)} + (1 - \alpha)Y,
   \]
   where \(Y^{(t)}\) is the label matrix at iteration \(t\), \(T\) is the transition matrix, \(Y\) is the initial label matrix, and \(\alpha \in [0,1]\) controls the balance between propagation and retention of initial labels.

\noindent\textbf{5. Convergence.} 
   The algorithm iterates until \(Y^{(t)}\) converges, meaning there is minimal change between successive iterations, ensuring the propagated labels are stable.

We use the off-the-shelf LableSpreading provided by scikit-learn~\cite{pedregosa2011scikit} in our implementation with their default settings, ensuring simple deployment to apply to different cases.

\appendixsection{Additional experts feedback}
Our experts provided feedback on additional aspects. One expert suggested that we directly display predictions and labels via text in each confusion matrix panel. ``\textit{It would be better if you show `label: waterbirds; prediction: landbirds' rather than `TN'}''. Another expert discussed the potential of applying our system to validate multi-classification models. Following our brief explanation of how \systemname{} produces results based on the model's predicted class without restrictions on the number of classes, the expert commented, ``\textit{Your system seems to be fully adaptable. And I would love to see this implementation in future work}.''

\appendixsection{Discussion and Future Work}
\label{appendix_sec:discussion}
% \vspace{.5mm}
\noindent\textbf{Potential risks introduced by XAI techniques.} \hspace{3mm}
In this work, we leverage state-of-the-art XAI techniques including GradCAM and Feature Inversion, where the first is one of the most widely-adopted attribution-based explanations for neural networks~\cite{rieger2020interpretations,arrieta2020explainable,linardatos2020explainable,ozturk2020automated} and the second is an advanced technique to visualize what a model is looking for~\cite{carter2019activation,hohman2019s,tjoa2020survey}. However, there has been continuing discussion on the reliability and trustworthiness of XAI techniques~\cite{adebayo2018sanity,ghorbani2019interpretation}, indicating the potential risks of using them. Even though both of them have been proven to retain a level of correctness through validations such as saliency checks~\cite{adebayo2018sanity}, we are aware of such risks and involve human auditions in our workflow to alleviate them. In the future, we plan to design more trustworthy model explanations by reasoning about the causality relationships in model decisions~\cite{pearl2009causality,halpern2020causes}, and continue to involve human-in-the-loop to mitigate pitfalls in AI-automated tools.

% \vspace{.5mm}
\noindent\textbf{Propagation approaches for the hypothetical spuriousness.} \hspace{3mm}
We adopt the Label Propagation method~\cite{zhou2003learning} to automatically spread the slices' spurious annotations from users to other slices as discussed in Sec. III-E (2). This is under the assumption that the distance between the slices corresponding to spurious correlations (e.g., water backgrounds) and the slices corresponding to core correlations (e.g., birds) is large. However, such assumptions can be invalid in other scenarios. For example, an image classification model trained on ImageNet~\cite{deng2009imagenet} includes various different classes such as ``husky dog'' and ``king penguin''. It is likely that two slices far away from each other are both corresponding to the core correlations. To address this, we plan to incorporate class similarities to the Label Propagation matrix to improve the precision of the generated hypothetical Spuriousness.

% \vspace{.5mm}
\noindent\textbf{Better guidance in data slice finding.} \hspace{3mm}
We include the model's performance indicators such as the classification accuracy and the average prediction confidence, as well as the Spuriousness probabilities in the Slice Table of our system to guide users to find interesting slices (refer to Fig. 2~\mybox{\textcolor{white}{B}}). Although the current guidance is proven to be effective in assisting users to detect slice issues, we plan to make further improvements. We aim to introduce new metrics calculated based on the model's performance, the slice pattern similarity, and the Spuriousness. By doing this, we will provide more informed guidance to our users by leveraging human and model's knowledge simultaneously, further improving the effectiveness of our model validation workflow.

% \vspace{.5mm}
\noindent\textbf{Extending to object detection and segmentation tasks.} \hspace{3mm}
Although our system is designed primarily for image classification, its core methodology---data slice generation and visualization techniques---can be extended to tasks such as object detection and segmentation. These extensions would require adjustments in slice definition and model performance metrics, but the attribution-based approach remains applicable. For instance, object detection models often provide region-specific attribution scores, which can be used to generate localized data slices. Similarly, segmentation tasks could leverage attribution maps to group pixels into coherent slices for validation. Future work will explore such extensions to broaden the applicability of our framework to a wider range of vision tasks and multimodal datasets.

% \vspace{.5mm}
\noindent\textbf{Scalability and computational efficiency improvements.} \hspace{3mm}
Our current system is optimized for handling large-scale datasets efficiently. Feature inversion, despite being an optimization-based method, is applied at the slice level rather than to individual instances, allowing for parallelized processing. Empirical results on CelebA (with $202k$ images) show that our feature inversion generates 50 slice visualizations in under 3 minutes using an Nvidia RTX 3090 GPU. Additionally, components such as UMAP, K-Means, and label propagation are designed to handle high-dimensional data at scale.
}
% % %%%%%%%%%%%%%%%%%%%%%%%%%%%%%%%%%%%%%%% \uparrow comment this out for main paper %%%%%%%%%%%%%%%%%%%%%%%%%%%%%%%%%%%%%%%

\bibliographystyle{IEEEtran}
\bibliography{Sections/References}

%%%%%%%%%%%%%%%%%%%%%%%%%%%%%%%%%%%%%%%% \downarrow comment this out for appendix %%%%%%%%%%%%%%%%%%%%%%%%%%%%%%%%%%%%%%%%%
% \vspace{-20pt}
\begin{IEEEbiography}[{\includegraphics[width=1in,height=1.25in,clip,keepaspectratio]{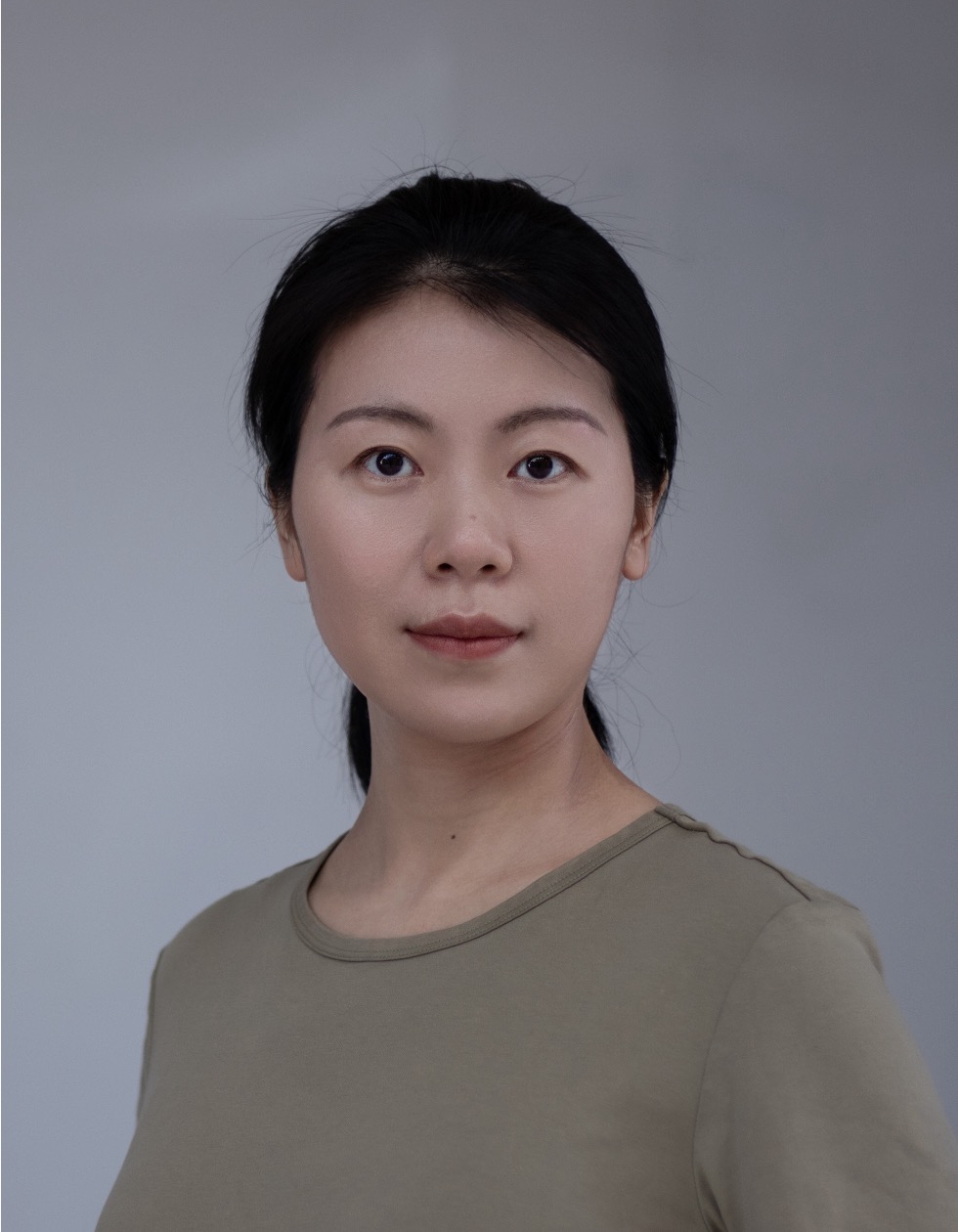}}]%
    {Xiwei Xuan} is a Ph.D. candidate in computer science at the University of California, Davis.
    Before UC Davis, she received her M.S. degree in electrical engineering in 2020 from Washington University in St. Louis.
    Her research spans computer vision, machine learning, and visual analytics, aiming to address the reliability, efficiency, and transparency of machine learning. She is particularly interested in improving data quality and the alignment between human and machine intelligence.
\end{IEEEbiography}
% \vspace{-20pt}
\begin{IEEEbiography}[{\includegraphics[width=1in,height=1.25in,clip,keepaspectratio]{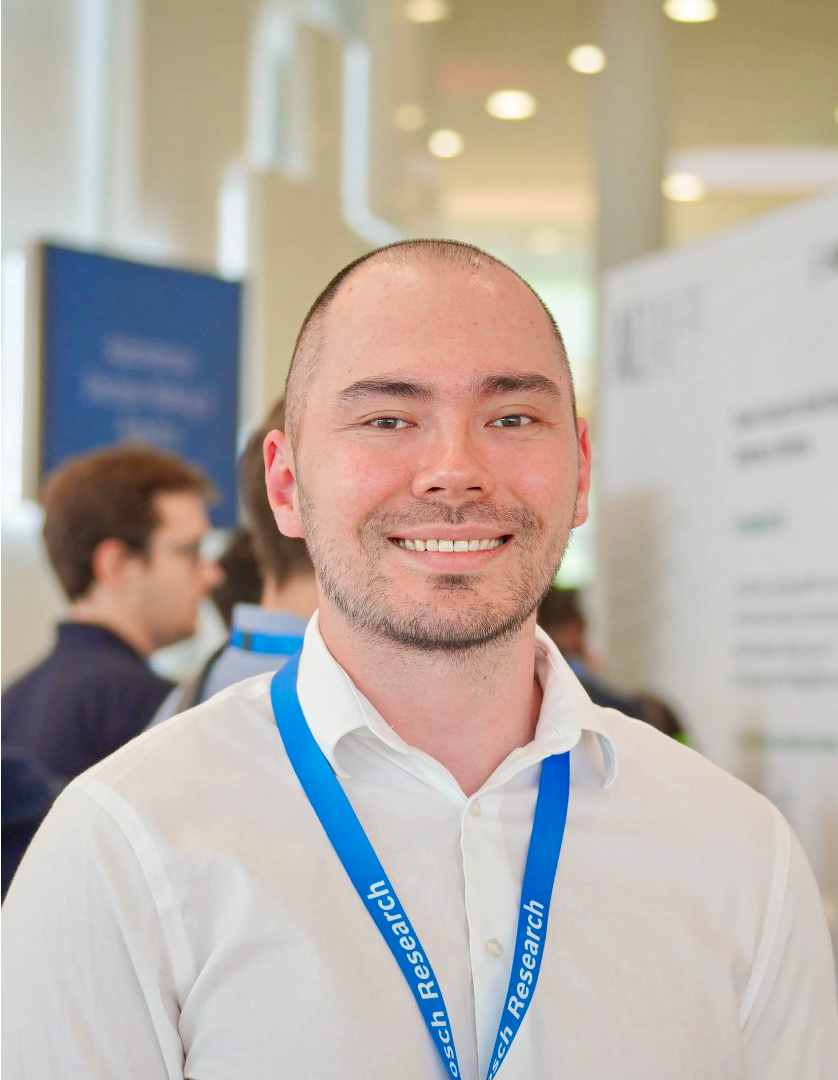}}]%
    {Jorge Piazentin Ono} is a Senior Research Scientist at Bosch. He earned his PhD in Computer Science from New York University and a Master’s degree from the University of Sao Paulo. His research interests include Human-Computer Interaction, Visual Analytics, Explainable AI, and Model Validation. Specifically, his work focuses on leveraging human domain knowledge and insights to improve data quality and model performance.
\end{IEEEbiography}
% \vspace{-20pt}
\begin{IEEEbiography}[{\includegraphics[width=1in,height=1.25in,clip,keepaspectratio]{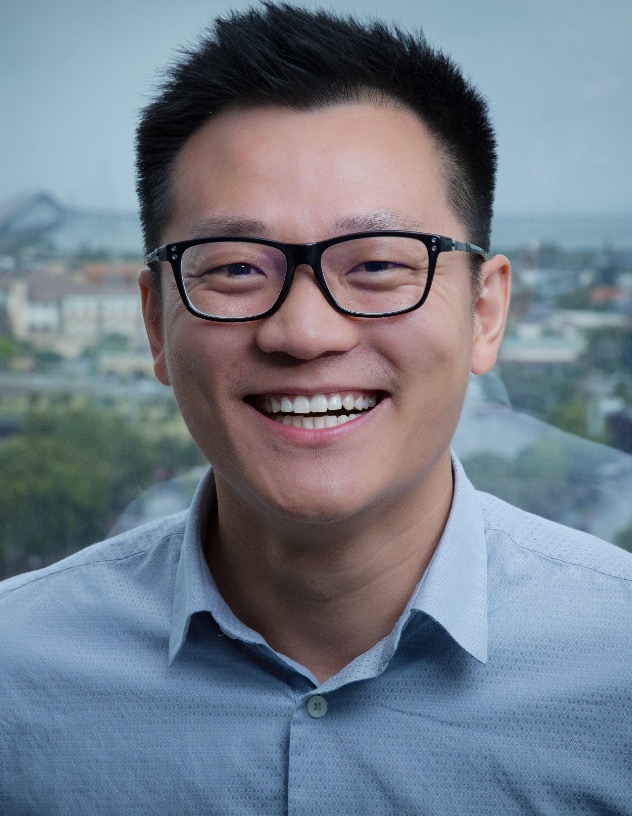}}]%
    {Liang Gou} is the Director of AI at Splunk, a part of Cisco. Previously, he served as a Senior Principal Research Scientist at Bosch, a Principal Research Scientist at Visa Research, and a Research Staff Member at IBM’s Almaden Research Center. Liang holds a Ph.D. in Information Science from Penn State University. His research interests focus on large language models (LLMs), agentic systems, and visual analytics.
\end{IEEEbiography}
% \vspace{-20pt}
\begin{IEEEbiography}[{\includegraphics[width=1in,height=1.25in,clip,keepaspectratio]{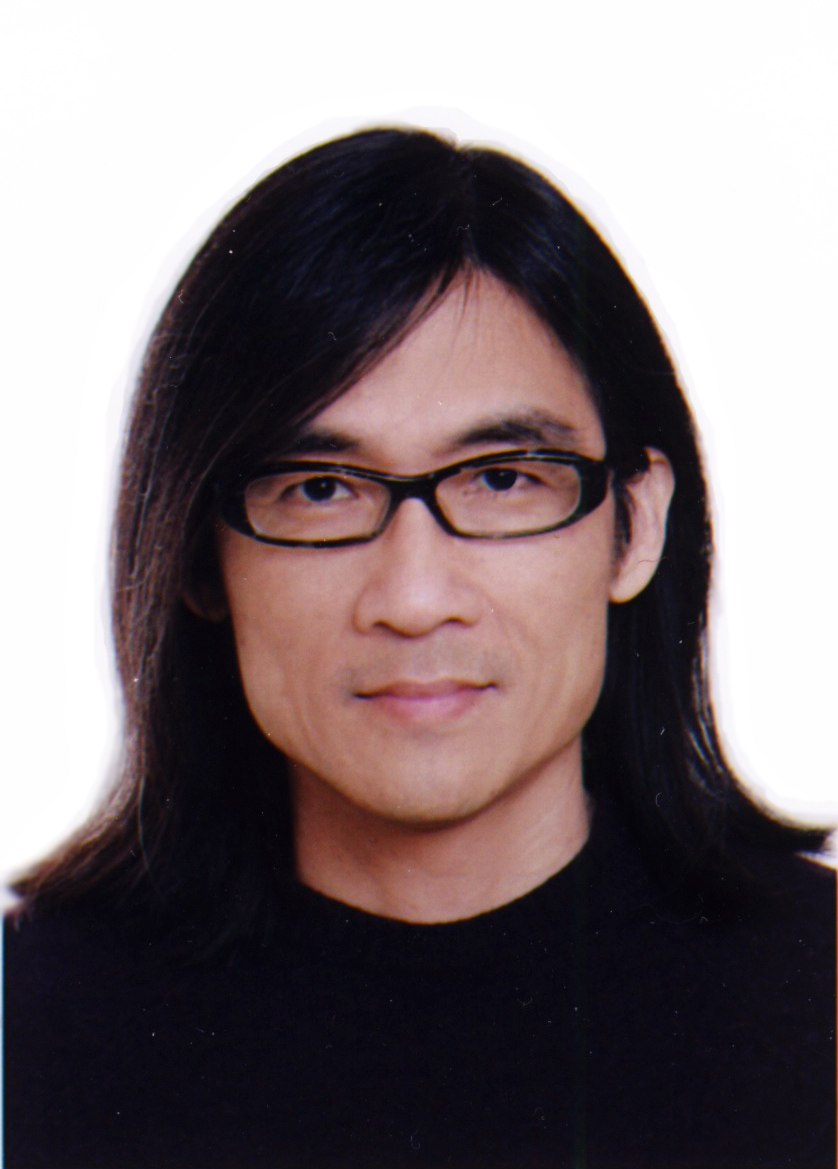}}]%
    {Kwan-Liu Ma} is a distinguished professor of computer science at the University of California, Davis. His research is in the intersection of data visualization, computer graphics, human-computer interaction, and high performance computing. For his significant research accomplishments, Ma received several recognitions including being elected as IEEE Fellow in 2012 and ACM Fellow in 2024, recipient of the IEEE VGTC Visualization Technical Achievement Award in 2013, and inducted to IEEE Visualization Academy in 2019.
\end{IEEEbiography}
% \vspace{-20pt}
\begin{IEEEbiography}[{\includegraphics[width=1in,height=1.25in,clip,keepaspectratio]{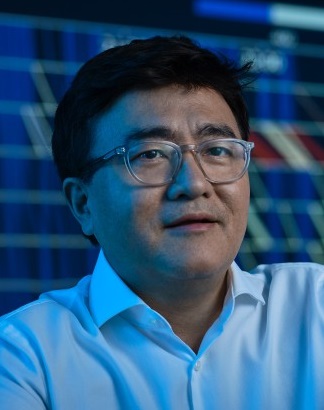}}]%
    {Liu Ren} is currently the Vice President and Chief Scientist of Salable and Assistive AI at Bosch Research North America and Bosch Center for AI (BCAI). He received his Ph.D. in Computer Science from the Computer Science Department at Carnegie Mellon University. His research focuses on AI, computer vision, and visual analytics, among other areas. He has been honored with multiple best paper honorable mention awards (2016,2022) and best paper awards (2018,2020) at IEEE Visualization conferences for his contributions to visual analytics \hspace*{1.15in}\mbox{and XAI}.
\end{IEEEbiography}
%%%%%%%%%%%%%%%%%%%%%%%%%%%%%%%%%%%%%%%%% \uparrow comment this out for appendix %%%%%%%%%%%%%%%%%%%%%%%%%%%%%%%%%%%%%%%%%
\vfill

\end{document}